\documentclass[journal]{IEEEtran}

\usepackage{graphicx}
\usepackage[caption=false,font=footnotesize]{subfig}
\usepackage{amsmath,amssymb,amsfonts}
\usepackage{algorithm}%
\usepackage{algorithmicx}%
\usepackage{algpseudocode}
\usepackage{adjustbox}  
\usepackage{multirow}
\usepackage{booktabs}
\usepackage{tabularx}
\usepackage{bm}

\ifCLASSINFOpdf
\else
\fi
\begin{document}

\title{GraphBEAT: Graph-based Bucketized Event Aggregation Over Time for Early Risk Stratification of Gastrointestinal Cancers}
\author{Leo Y. Li-Han,
        Chris Varghese,
        Elizabeth B. Habermann, 
        Mark J. Truty,
        William R.G. Perry, \\ 
        Cornelius A. Thiels, and Hojjat Salehinejad, \IEEEmembership{Senior Member, IEEE}
\thanks{Leo Y. Li-Han, 
        Chris Varghese,
        Mark J. Truty,
        William R.G. Perry, and Cornelius A. Thiels are with the Department of Surgery, Mayo Clinic, Rochester, MN, USA .}
\thanks{Elizabeth B. Habermann is with the Kern Center for the Science of Health Care Delivery and the Department of Surgery, Mayo Clinic, Rochester, MN, USA.}
\thanks{Hojjat Salehinejad is with the Kern Center for the Science of Health Care Delivery and the Department of Artificial Intelligence and Informatics, Mayo Clinic, Rochester, MN, USA.}}

\maketitle

\begin{abstract}
Longitudinal electronic health records (EHRs) contain complementary diagnostic and laboratory information about patient health. However, irregularity, sparsity, heterogeneity, and incompleteness in documentation complicate EHR modeling. To address the problem, we introduce GraphBEAT (Graph-based Bucketized Event Aggregation over Time), a temporal graph representation learning framework that groups clinical events into time buckets and transforms observed features into patient-specific, temporally directed graphs to enable early risk stratification for gastrointestinal cancers. Each node in an EHR graph encodes feature identity, observed value, and temporal position of clinical events. A context-aware, time-decay message-passing mechanism models transitions between successive buckets, followed by multi-head attention pooling to generate patient-level representations. GraphBEAT was trained and evaluated on multi-site longitudinal diagnosis codes and laboratory tests to predict pancreatic, liver, and intestinal cancer at 12-, 24-, and 36-month diagnostic lead times. In all experimental setups, GraphBEAT consistently outperformed baseline graph neural networks (GraphSAGE, GAT, and GCN) and sequence modeling approaches (Transformer and long short-term memory [LSTM]). GraphBEAT also required over 250$\times$ fewer floating-point operations per patient and over 20$\times$ fewer parameters than the Transformer and LSTM baselines. These results support GraphBEAT as an accurate and computationally efficient approach for cancer risk stratification from sparse, heterogeneous longitudinal health records.
\end{abstract}


%
\IEEEpeerreviewmaketitle

\section{Introduction}

\IEEEPARstart{L}{ongitudinal} electronic health records (EHRs) contain extensive and complementary information on the evolution of patient health, including diagnostic trajectories, laboratory measurements, healthcare utilization patterns, and other routinely collected clinical data~\cite{amirahmadi2023deep}. Diagnosis codes provide standardized documentation of diseases, symptoms, and comorbidities, while laboratory measurements record quantitative observations of physiological state. An integrated assessment of heterogeneous EHR data could reveal subtle interactions among symptoms, conditions, and measurements that may not be apparent from either modality alone, thereby supporting clinical tasks such as disease-onset prediction, patient triage, and early risk stratification~\cite{amirahmadi2023deep, li2020behrt}. 

This opportunity is particularly relevant to gastrointestinal malignancies that may present at a younger age or remain clinically occult until advanced stages~\cite{us2019screening, villanueva2019hepatocellular, raghav2013smallbowel, dekker2019colorectal}. For instance, population-wide screening for pancreatic cancer is economically impractical given its low prevalence and potential harms of invasive screening~\cite{us2019screening, dbouk2022multicenter}. Meanwhile, established surveillance programs are limited to selected high-risk subgroups that account for only a minority of future cases~\cite{sharma2018model, goggins2020management, aslanian2020aga}. Recent studies suggest that deep learning models can enhance pancreatic cancer risk prediction from routinely collected longitudinal clinical records ~\cite{jia2023pancreatic, mavromatis2026prime, placido2023deep, zheng2025pancreatic, varghese2026digitally, park2025enhancing}. Although screening pathways may differ between pancreatic, liver, and intestinal cancers, a scalable modeling method that identifies a manageable high-risk subgroup from routine EHR data could complement existing methods by prioritizing patients for further diagnostic assessment~\cite{varghese2026digitally}.

Modeling longitudinal EHR data is nevertheless challenging because clinical encounters/events are irregular and heterogeneous, and the records may be repeated or inaccurate~\cite{amirahmadi2023deep}. Diagnosis codes are categorical and sparse, and the accuracy and completeness vary depending on coding and billing practices~\cite{hripcsak2013next}. Laboratory measurements are continuous, have category-specific value distributions, and are collected selectively according to clinical need~\cite{aarsand2018eubivas}. Patients also differ in their healthcare utilization patterns, observation duration, and record completeness. Therefore, an effective and generalizable modeling approach should preserve event identity, value, and timing in data representation while learning clinically meaningful dependencies across heterogeneous measurements over long temporal horizons~\cite{amirahmadi2023deep}.

\begin{figure*}[!ht]
    \centering
    \includegraphics[width=0.99\textwidth]{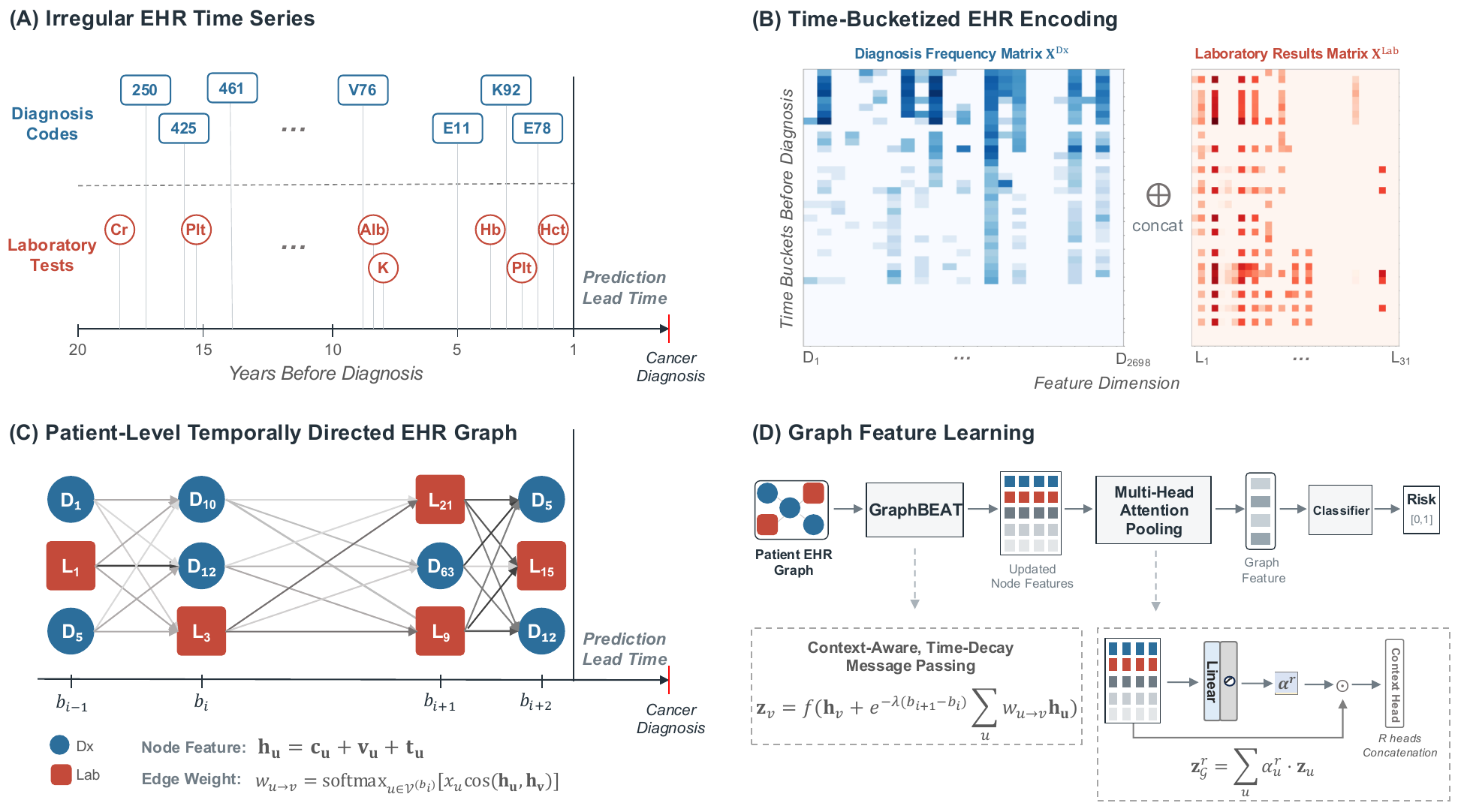}
    \caption{The graph representation learning framework for cancer risk prediction from heterogeneous EHR time series. \textbf{(A)} An example of irregular diagnosis and laboratory trajectories for one patient. The task is to predict the risk of pancreatic, liver, and intestinal cancer using up to 20 years of clinical history, with diagnostic lead times of 12, 24, and 36 months. \textbf{(B)} Heatmap of the matrix representations of the diagnostic codes (blue) and laboratory results (red) for the patient. Two matrices are concatenated along the feature dimension to form the heterogeneous EHR matrix. \textbf{(C)} The patient-specific temporally directed EHR graph constructed from the matrix representation by retaining observed clinical events. The node feature is determined by clinical feature identity, observed values, and temporal position, while the edge weights are dynamically calculated from the context of the transition. \textbf{(D)} The end-to-end patient-level graph learning for cancer risk prediction. The context-aware, time-decay message passing updates node features, while the multi-head attention feature pooling aggregates a varying number of node features into a fixed-length graph feature vector for downstream classification.}
    \label{fig1_architecture}
\end{figure*}

Existing methods have primarily represented clinical histories as temporal token sequences or graphs~\cite{placido2023deep, choi2017gram}. Sequence models capture temporal dependencies but often encode repeated or miscoded events as separate tokens and leave relationships among heterogeneous clinical variables implicit. Graph-based models make relational structure explicit, but many organize time through event hierarchies by involving external ontologies, knowledge graphs, and auxiliary temporal encoders~\cite{choi2017gram, choi2020gct, jiang2024graphcare, chen2024trans, kauffman2025infehr, zhang2026ehrformer}. Relatively less attention has been given to a patient-specific graph in which heterogeneous clinical observations share a common temporal topology and allow direct feature interaction across different time intervals. To approach this, our previous proof-of-concept study, GraD-IBD~\cite{lihan2026gradibd}, investigated time-bucketized graphs for inflammatory bowel disease detection using diagnosis histories alone. Its initial node representations used categorical diagnosis embeddings, while diagnosis frequencies and elapsed time informed temporally weighted message passing.

Following this philosophy, we introduce GraphBEAT, the \textit{Graph}-based \textit{B}ucketized \textit{E}vent \textit{A}ggregation over \textit{T}ime framework for irregular longitudinal EHR modeling for risk stratification of gastrointestinal cancers. GraphBEAT extends data representation to heterogeneous EHR trajectories (i.e., diagnosis and laboratory) by jointly encoding clinical feature identity, observed value, and temporal position in a shared latent space, enabling direct heterogeneous feature extraction through a common temporal graph topology. The framework retains the context-aware, time-decay message-passing principle while introducing adaptive multi-head attention pooling to integrate evidence across the complete clinical history. 

We evaluate GraphBEAT for pancreatic, liver, and intestinal cancer risk prediction at three diagnostic lead times using geographic holdout experiments across four Mayo Clinic care sites. The main contributions of the work include:
\newpage

\begin{itemize}
\item \textbf{Unified representation of irregular EHR data}: \\We introduce a clinically motivated time-bucketized encoding paradigm that transforms heterogeneous diagnosis codes and laboratory measurements into a patient-specific temporally directed EHR graph, preserving event identity, observation, and temporal position.

\item \textbf{Context-aware, time-decay graph learning}: \\We develop a three-component node representation and a message passing mechanism that encodes event context, source–target transition relevance, and temporal recency into the graph representation learning. An adaptive multi-head attention feature pooling subsequently integrates updated node representations over time into a fixed-dimensional patient representation.

\item \textbf{Multi-site and multi-cancer evaluation}: \\We evaluate the framework across four geographically and operationally distinct care settings, three cancer types, and three diagnostic lead times using geographic internal-external validation, and compare it with established graph- and sequence-based architectures.
\end{itemize}

\section{Related Work}

\subsection{Sequence Models for EHR Representation Learning}
As longitudinal EHR data can be naturally represented as sequences of clinical events, sequence modeling approaches have long been a primary methodology in data analysis. Early modeling efforts focused on recurrent neural networks (RNNs), such as Long Short-Term Memory (LSTM) and Gated Recurrent Units (GRU)~\cite{hochreiter1997long, chung2014empirical, salehinejad2017recent, choi2016doctor}. With the striking success in natural language processing, recent sequence approaches in EHR modeling have mainly centered on Transformer architectures~\cite{vaswani2017attention}. Li~\textit{et al.} trained a Transformer Encoder model to jointly embed diagnoses, age, visit segments, and event positions to learn longitudinal patient representations for predicting future diseases~\cite{li2020behrt}. In pancreatic cancer risk prediction, Placido~\textit{et al.} demonstrated that Transformer-based models can extract predictive information from the ordering and timing of diagnosis trajectories~\cite{placido2023deep}. Subsequently, different customized Transformer architectures were developed to model heterogeneous diagnosis, laboratory, and medication trajectories for improved performance~\cite{zheng2025pancreatic, varghese2026digitally}. Meanwhile, feature representations generated by pretrained language models were also investigated for pancreatic cancer risk prediction~\cite{park2025enhancing}. Together, these studies establish the value of longitudinal and multimodal clinical information. Nevertheless, sequence models typically treat each clinical event as a separate token. Repeated or miscoded documentation in diagnoses can therefore lengthen input trajectories, affect feature extraction, and increase computational burden, while temporal relationships among clinical events remain implicit in the input representation. Furthermore, as multimodal measurements differ in representation and sampling process, joint modeling may also require modality-specific encoders and feature fusion modules~\cite{xu2021mufasa}.

\subsection{Graph-based EHR Representation Learning}
Graph representation learning offers a complementary methodological solution by making relationships among clinical events, measurement patterns, and temporal states explicit. Consequently, graph neural networks (GNNs) can model both inter-variable and inter-temporal relationships that are difficult to specify directly in conventional sequence representations~\cite{jin2024survey, choudhary2024graph, lihan2026gradibd}. 
Cheng~\textit{et al.} demonstrated a general series-as-graph paradigm in which time series subsequences and their evolving relationships are learned through multiple attention mechanisms~\cite{cheng2023time2graph}. In EHR modeling, Choi~\textit{et al.} incorporated medical-ontology structure into diagnosis representations to learn latent relationships among clinical concepts~\cite{choi2017gram, choi2020gct}. More recent GNN models have increasingly incorporated temporal structure into patient graphs. For example, Chen~\textit{et al.} represented visits and medical events within patient-specific temporal multimodal graphs and integrated temporal edge features into graph transformation~\cite{chen2024trans}. Kauffman~\textit{et al.} converted longitudinal EHRs into temporal graphs for probabilistic clinical inference~\cite{kauffman2025infehr}, while Zhang~\textit{et al.} constructed patient-specific graphs containing timestamped visit nodes and multimodal medical-event nodes for next-visit diagnosis prediction~\cite{zhang2026ehrformer}. All the studies demonstrate the growing utility of graph representation learning in jointly modeling structural and temporal information. Despite the progress, many existing EHR graph models emphasize ontology-enriched concept relations~\cite{choi2017gram}, hierarchical structures~\cite{choi2020gct}, imported clinical knowledge~\cite{jiang2024graphcare}, and additional temporal encoders~\cite{chen2024trans}. A relatively underexplored design is to represent the entire clinical history as a patient-specific, layered graph in which heterogeneous data share one temporal topology and interact directly across clinical intervals. 

To bridge the gap, we introduce GraphBEAT, which represents diagnosis codes and laboratory measurements within a unified, time-bucketized, temporally directed graph for longitudinal cancer risk prediction.

\section{Methods}
We formulate per-patient cancer risk prediction as a graph-level binary classification task, which comprises three main steps. First, irregular diagnostic codes and laboratory results are grouped into fixed-length time buckets and encoded as structured matrices. Second, the matrices are converted into temporally directed EHR graphs, such that node features and edge connections preserve event identity, observed value, and temporal order. Third, a context-aware, time-decay message-passing mechanism models transitions between successive clinical states, after which adaptive multi-head attention aggregation produces a graph-level feature representation for classification. Fig.~\ref{fig1_architecture} illustrates the overall GraphBEAT framework. 

\subsection{Irregular EHR Data Encoding}
For a given patient, the observation period of EHR data is first discretized into $T$ fixed-length time buckets, each spanning $\tau$ days. Let $\mathcal{D}=\{d_1,d_2,\ldots,d_{N_{\mathrm{D}}}\}$ denote a vocabulary of $N_{\mathrm{D}}$ diagnosis codes in the dataset. As such, the patient's diagnosis history can be encoded as a matrix $\mathbf{X}^{\mathrm{Dx}}\in\mathbb{R}^{N_{\mathrm{D}}\times T}$, where each row represents a diagnosis code, each column represents a time bucket, and each element $x^{\mathrm{Dx}}_{c,t}$ records the number of times diagnosis code $d_c$ appears in bucket $t$. Repeated occurrences of the same diagnosis within a bucket are summarized by their frequency, whereas unobserved diagnoses are represented by zero.

Similarly, let $\mathcal{L}=\{\ell_1,\ell_2,\ldots,\ell_{N_{\mathrm{L}}}\}$ be the set of laboratory test categories in the dataset with a vocabulary size of $N_{\mathrm{L}}$. 
Using the same temporal buckets, the patient's laboratory history is encoded as $\mathbf{X}^{\mathrm{Lab}}\in\mathbb{R}^{N_{\mathrm{L}}\times T}$. Each element $x^{\mathrm{Lab}}_{j,t}$ contains the mean of the available measurements for the laboratory test $\ell_j\in\mathcal{L}$ in bucket $t$, whereas unobserved laboratory tests are stored as zero.

The two resulting matrices are then concatenated along the feature dimension to form the heterogeneous encoding matrix $\mathbf{X}=[\mathbf{X}^{\mathrm{Dx}};\mathbf{X}^{\mathrm{Lab}}] \in\mathbb{R}^{N\times T}$, where  $N=N_{\mathrm{D}}+N_{\mathrm{L}}$ and $\mathcal{F}=\mathcal{D} \cup \mathcal{L}$ represents the combined clinical feature vocabulary. Accordingly, the row, column, and value of each entry in $\mathbf{X}$ encode the feature identity, temporal position, and diagnosis frequency/laboratory measurement, respectively (Figure~\ref{fig1_architecture}B).

\subsection{Patient-Level Temporally Directed EHR Graph}
The encoding matrix $\mathbf{X}$ is further converted into a patient-specific, layered directed acyclic graph $\mathcal{G}=(\mathcal{V},\mathcal{E})$ that retains only observed features, where $\mathcal{V}$ and $\mathcal{E}$ represent the node and edge sets in the graph, respectively.

Let $\mathcal{B}=(b_1,b_2,\ldots,b_B)$ denote an ordered set of non-empty time buckets (columns) in $\mathbf{X}$, indexed such that $b_1<b_2<\cdots<b_B$ where $B\leq T$. For each bucket $b_i\in\mathcal{B}$, a node subset $\mathcal{V}^{(b_i)}$ is created with one node for every unique clinical feature $c\in\mathcal{F}$ observed in that bucket. Binary feature observation masks were also constructed to mark observed features and represent unobserved entries by node absence. Each node is associated with a stack of three attributes $(c, x_{c, b_i}, b_i)$, corresponding to its feature identity, bucket-level value, and temporal position, respectively. These attributes are then transformed into learnable node representations in the same latent space, which will be elaborated in the next subsection. The complete node set is the union of node subsets $\mathcal{V}=\bigcup_{i=1}^{B}\mathcal{V}^{(b_i)}$. The number of nodes $|\mathcal{V}|$ equals the number of observed feature–bucket pairs and thereby can vary across patient graphs.

Directed edges are constructed from every node in bucket $b_i$ to every node in the subsequent bucket $b_{i+1}$, defined as $\mathcal{E} = \bigcup_{i=1}^{B-1} \left( \mathcal{V}^{(b_i)} \times \mathcal{V}^{(b_{i+1})} \right)$, where $\times$ is the Cartesian product. Thus, this shared topology allows direct within- and cross-modality message passing, without learning separate modality-specific representations. No edges are introduced within a bucket because events occurring over a short period are collectively treated as a patient snapshot or clinical state rather than as separate events with strong temporal dependencies, which can further simplify the architecture and computation.

\subsection{Learning on the Graph}
The constructed graph topology defines the possible transitions across different data sources, while the strength of each transition is computed dynamically from the source and target node representations. For notational simplicity, here we define $v=v_{d,b_{i+1}}$ to represent a target node in bucket $b_{i+1}$ and $u=v_{c,b_{i}}$ to denote one of its source nodes in bucket $b_{i}$.

\subsubsection{Node Feature Embedding}
As each node contains three complementary attributes, i.e., feature identity, bucket-level value, and temporal position, the node feature vector is correspondingly constructed from three embedding components:

\begin{itemize}
    \item The \textbf{feature identity} of node $u$ is represented by $\mathbf{c}_u = \mathbf{C}_{c,:} \in\mathbb{R}^{d_{\mathrm{node}}}$, where $\mathbf{C}\in\mathbb{R}^{N\times d_{\mathrm{node}}}$ is a learnable embedding matrix shared across all patients and $d_{\mathrm{node}}$ is the node feature dimension. Here, diagnosis and laboratory categories are embedded in a common latent space defined by $\mathbf{C}$.

    \item The \textbf{bucket-level value} $x_{u}$ is projected into the same $d_{\mathrm{node}}$-dimensional space through a learnable linear mapping $\mathbf{v}_u = f_{\mathrm{val}}(x_{u}) : \mathbb{R} \rightarrow \mathbb{R}^{d_{\mathrm{node}}}$, where $x_{u}$ corresponds to the value of associated with node $u$. The same mapping is shared across patients and both data sources.

    \item The \textbf{temporal encoding} of all nodes in bucket $b_i$, denoted as $g_{\mathrm{pos}}(b_i) = \mathbf{t}_{b_i} \in \mathbb{R}^{d_{\mathrm{node}}}$, is represented using a sinusoidal encoding function given in~\cite{vaswani2017attention}.
\end{itemize}

Therefore, the feature representation of node $u$ is the element-wise addition of the three embedding vectors in the same latent space, which is given by

\begin{equation}
    \mathbf{h}_u = \mathbf{c}_u + \mathbf{v}_u + \mathbf{t}_{b_i}
    \in \mathbb{R}^{d_{\mathrm{node}}}.
    \label{eq:node_embedding}
\end{equation}

Here, nodes representing the same clinical feature share the same category embedding but can have distinct feature representations because of differences in their observed values and temporal positions.

\subsubsection{Dynamic Edge Weighting}
For an edge from node $u$ to node $v$, denoted as $e_{u\rightarrow v}$, the strength of the connection, i.e., edge weight, is dynamically computed in the forward pass based on two factors: 

\begin{itemize}
    \item The \textbf{relevance of the transition} is quantified using the cosine similarity between the source and target node features, denoted $\operatorname{cos}(\mathbf{h}_u, \mathbf{h}_v) = \frac{\mathbf{h}_u \cdot \mathbf{h}_v}{\lVert \mathbf{h}_u\rVert \lVert \mathbf{h}_v\rVert}$, where $\cdot$ denotes the dot-product and $\lVert \cdot \rVert$ represents the L2-norm.
    
    \item A \textbf{data-driven scaling term} that is defined by the source node value $x_{u}$, representing the current (source) status. 
\end{itemize}

Subsequently, the softmax function is applied to the raw weights across all incoming edges from source nodes $u\in\mathcal{V}^{(b_i)}$ to the target node $v\in\mathcal{V}^{(b_{i+1})}$ for weight normalization. The overall edge weighting process is therefore formulated as

\begin{equation}
    w_{u\rightarrow v} =
    \frac{ 
        \exp \left[   x_{u} \operatorname{cos}(\mathbf{h}_u, \mathbf{h}_v) \right]}{ 
        \displaystyle
        \sum_{q \in \mathcal{V}^{(b_i)}}
        \exp \left[  x_{q} \operatorname{cos}(\mathbf{h}_q, \mathbf{h}_v) \right]}.
    \label{eq:edge_weight}
\end{equation}

\begin{algorithm*}[!t]
\small
\caption{Context-aware, time-decay graph feature learning}
\label{alg:message_passing}
\begin{algorithmic}[1]
    \Statex \textbf{Input:}
    Patient graph $\mathcal{G}=(\mathcal{V},\mathcal{E})$ with temporally ordered buckets $\{\mathcal{V}^{(b_i)}\}_{i=1}^{B}$; 
    node attributes $\{(c_v,x_v,b_v)\}_{v\in\mathcal{V}}$; 
    category embedding matrix $\mathbf{C}$; value mapping $f_{\mathrm{val}}$; 
    temporal encoding $g_{\mathrm{pos}}$; search depth $K$; message update parameters $\mathbf{W}^k_\mathbf{m}, \mathbf{b}^k_\mathbf{m}, \forall k\in \{1,\ldots,K\}$; time-decay parameter $\lambda$; non-linearity $\sigma$
    
    \Statex \textbf{Output:} Node feature matrix $\mathbf{H}_{\mathcal{G}} $
    \State
    $\mathbf{h}^{0}_v \gets 
    \mathbf{C}_{c_v,:} + f_{\mathrm{val}}(x_v) + g_{\mathrm{pos}}(b_v),
    \quad \forall v\in\mathcal{V}$;
    \State
    $w_{u\rightarrow v} \gets
    \operatorname{softmax}_{u\in\mathcal{V}^{(b_{i})}} 
       \left[ x_u \operatorname{cos}(\mathbf{h}^{0}_u, \mathbf{h}^{0}_v) \right],
    \quad \forall u\in\mathcal{V}^{(b_{i})}, v\in\mathcal{V}^{(b_{i+1})}, i=1,\ldots,B-1$;
    
    \For{$k=1,\ldots,K$}
        \State
        $\mathbf{h}^{k}_v \gets \sigma\!\left( \mathbf{W}^{k}_{\mathbf m}\mathbf{h}^{k-1}_v + \mathbf{b}^{k}_{\mathbf m} \right), \quad \forall v\in\mathcal{V}^{(b_1)}$;
        
        \For{$i=1,\ldots,B-1$}    
            \State
            $\widetilde{\mathbf{m}}^k_v \gets
            \exp\!\left[-\lambda(b_{i+1}-b_i)\right] 
            \displaystyle 
            \sum_{u\in\mathcal{V}^{(b_i)}}
            w_{u\rightarrow v}
            \mathbf{h}^{k-1}_u,
            \quad \forall v\in\mathcal{V}^{(b_{i+1})}$;
        
            \State
            $\mathbf{h}^k_v \gets
            \sigma \left( \mathbf{W}_{\mathbf{m}}^{k} 
                   \left( \mathbf{h}^{k-1}_v + \widetilde{\mathbf{m}}^{k}_v \right) + \mathbf{b}^{k}_{\mathbf{m}} \right), \quad \forall v\in\mathcal{V}^{(b_{i+1})}$;
        \EndFor
    \EndFor
    \State
    $\mathbf{z}_{v} \gets \mathbf{h}^{K}_v, \quad \forall v\in \mathcal{V}$
    \State \Return $\mathbf{H}_{\mathcal{G}} \gets \left[ \mathbf{z}_{v} \right]_{v\in \mathcal{V}}$
\end{algorithmic}
\end{algorithm*}

\subsubsection{Context-Aware, Time-Decay Message Passing}
Algorithm~\ref{alg:message_passing} presents the workflow of the proposed message-passing mechanism, where the target node representation is iteratively updated using feature vectors from its incoming nodes. Let $K$ denote the message-passing search depth that determines the temporal receptive field for feature updates. Each layer propagates information across one directed edge between consecutive non-empty buckets. After $K$ layers, a node may incorporate information from at most $K$ preceding buckets through directed multi-hop paths. 
In our experiments, $K$ is set to $1$ after hyperparameter tuning. Thus, we omit the search depth-specific superscripts in the following expressions, while retaining them in Algorithm~\ref{alg:message_passing} . 

First, messages from all source nodes $u\in\mathcal{V}^{(b_i)}$ associated with the target node $v$ are aggregated through a weighted sum according to the normalized edge weights. To account for the irregular temporal interval between two consecutive buckets, the aggregated message is modulated by an exponential time-decay factor determined by the time difference $b_{i+1}-b_i$, yielding the decayed message as
\begin{equation}
    \widetilde{\mathbf{m}}_v =
    \exp\left[ -\lambda(b_{i+1}-b_i) \right] 
    \sum_{u\in\mathcal{V}^{(b_i)}} w_{u\rightarrow v} \mathbf{h}_u \in \mathbb{R}^{d_{\mathrm{node}}},
    \label{eq:time_decay_message}
\end{equation}
where $\lambda \geq 0$ is a learnable parameter that controls the decay strength and is optimized with the model parameters. 

The target node feature is updated by element-wise addition of its current feature vector and the time-decayed neighborhood message, followed by a fully connected transformation, parameterized by $\mathbf{W_m}\in \mathbb{R}^{d_{\mathrm{graph}}\times d_{\mathrm{node}}}$ and $\mathbf{b_m} \in \mathbb{R}^{d_{\mathrm{graph}}}$, with the Gaussian Error Linear Unit (GELU) activation function~\cite{hendrycks2016gelu}, to generate the updated feature representation of the target node $v$ as
\begin{equation}
    \mathbf{z}_v = \operatorname{GELU} \left(
        \mathbf{W_m} \left( \mathbf{h}_v + \widetilde{\mathbf{m}}_v \right) + \mathbf{b_m}
    \right) \in \mathbb{R}^{d_{\mathrm{graph}}},
    \label{eq:node_update}
\end{equation}
where $d_{\mathrm{graph}}$ represents the graph embedding dimension. 

In this process, the self-representation $\mathbf{h}_v$ preserves information about the target feature, whereas the message $\widetilde{\mathbf{m}}_v$ incorporates contextually weighted and temporally modulated information from the preceding bucket. For nodes in the first bucket, which have no incoming edges, the aggregated message is set to zero, and the same linear transformation and activation are applied to their self-representations. Therefore, this mechanism passes more information from recent, relevant, and important clinical events to the target node, aligning with the clinical reasoning and thereby being termed context-aware, time-decay message passing. The final node representations form the graph feature matrix, which is given by
\begin{equation}
    \mathbf{H}_{\mathcal{G}} = \left[ \mathbf{z}_v \right]_{v\in\mathcal{V}}
    \in \mathbb{R}^{|\mathcal{V}|\times d_{\mathrm{graph}}}.
    \label{eq:graph_feature_matrix}
\end{equation}

\subsection{Feature Pooling and Classification}
Because the number of observed nodes varies across EHR graphs (patients), we introduce a multi-head attention pooling layer to transform the updated node feature matrix $\mathbf{H}_{\mathcal{G}}$ into a fixed-length graph feature vector $\mathbf{z}_\mathcal{G}$. 

Let $R$ denote the number of attention heads. Each attention head learns an independent scoring function that assigns an importance coefficient to weigh every node in the graph. Since different node features encode the patient's conditions at varying time points, this pooling layer adaptively integrates long-range evidence through graph-level attention over temporally encoded node representations. The aggregated graph-level feature representation from attention head $r$, denoted as $\mathbf{z}^{r}_{\mathcal{G}}$, can be expressed as
\begin{align}
    \bm{\alpha}_r &= \operatorname{softmax}(\mathbf{H}_{\mathcal{G}}\mathbf{w}_r) \in \mathbb{R}^{|\mathcal{V}|} \\
    \mathbf{z}^{r}_{\mathcal{G}} &= \bm{\alpha}_r^\top \mathbf{H}_{\mathcal{G}}
    \in \mathbb{R}^{d_{\mathrm{graph}}}
    \label{eq:graph_embedding}
\end{align}
where $\mathbf{w}_r\in \mathbb{R}^{d_{\mathrm{graph}}}$ is the learnable parameter to calculate attention scores for all nodes from attention head $r$. 

Finally, the pooled graph feature vectors from all attention heads are concatenated, forming the graph feature representation $\mathbf{z}_\mathcal{G}=[\mathbf{z}_\mathcal{G}^1; \ldots; \mathbf{z}_\mathcal{G}^r; \ldots; \mathbf{z}_\mathcal{G}^R] \in \mathbb{R}^{R d_{\mathrm{graph}}}$, which is passed to a multilayer perceptron classifier with two hidden-layer to produce the predicted cancer probability and binary outcomes.

\section{Experiments}
\subsection{Data}

This study involves analysis of multi-site, de-identified EHR data via the Mayo Clinic Platform~\cite{yu2026accelerating}. The Platform is HIPAA-compliant. Any data beyond what is reported in the manuscript, including but not limited to the raw EHR data, cannot be shared or released due to the parameters of the expert determination to maintain the data de-identification. This study was approved by the Mayo Clinic Institutional Review Board. 

The source population included patients receiving care at three tertiary and quaternary Mayo Clinic medical centers (Rochester, Arizona, and Florida) and at the Mayo Clinic Health System (MCHS) that comprises 17 community hospitals. The included data contained time-stamped International Classification of Diseases (ICD) diagnosis codes (versions 9 and 10) and measurements from 31 prespecified routine blood tests, spanning up to 20 years of clinical history before the index date, ranging from 1985 to 2024. All ICD-9 and ICD-10 diagnosis codes are truncated to the first 3 characters based on modeling conventions~\cite{li2020behrt, placido2023deep, zheng2025pancreatic}, while lab measurements are normalized per test category based on their corresponding statistics from the training sets.

We defined three graph-level binary classification tasks for pancreatic cancer, liver cancer, and intestinal cancer risk predictions, respectively. Cases were identified using cancer-specific ICD-9 or ICD-10 codes, with the first qualifying diagnosis defining the cancer index date. Non-cancer controls had no qualifying diagnosis code for any gastrointestinal cancer category, including pancreatic, esophageal, gastric, gallbladder or extrahepatic bile duct, liver or intrahepatic bile duct, small- or large-intestinal cancer. The case and control groups were selected in an age- and sex-matched manner. To reduce the impact of occult malignancy near the end of follow-up, the index date for the control group was defined as one year before the last recorded encounter, following the method of~\cite{placido2023deep, varghese2026digitally}. Patients with less than 3 years of clinical history or fewer than 3 qualifying laboratory observations were excluded. Table~\ref{tab:cohort} details the diagnosis code-based cohort definition criteria and included laboratory tests.

\begin{table}[!t]
    \centering
    \caption{Cohort definition criteria and included laboratory tests.}
    \label{tab:cohort}
    \begin{adjustbox}{max width=1\columnwidth,center}
    \begin{tabular}{@{}ll@{}}
        \toprule
        Cohort component & Definition diagnosis codes or included lab tests \\ 
        \midrule
        Pancreatic cancer & 157, C25 \\
        \addlinespace[3pt]
        Liver cancer & 155, C22 \\
        \addlinespace[3pt]
        Intestinal cancer & 152, 153, 154, C17, C18, C19, C20, C21 \\
        \addlinespace[3pt]
        \begin{tabular}[c]{@{}l@{}}Control \\ (excluded codes)\\~\end{tabular} & 
        \begin{tabular}[c]{@{}l@{}}150, 151, 152, 153, 154, 155, 156, 157, 230\\ 
        C15, C16, C17, C18, C19, C20, C21, C22, C23, C24, C25\\D00, D01, C7A\end{tabular} \\
        \addlinespace[4pt]
        Laboratory tests & 
        \begin{tabular}[t]{@{}l@{}}
        Albumin, Alkaline Phosphatase, Alanine Aminotransferase,\\
        Aspartate Aminotransferase, Gamma Glutamyl Transferase,\\
        Total Bilirubin, Amylase, Lipase, Blood Urea Nitrogen,\\
        Creatinine, Calcium, Chloride, Magnesium, Potassium,\\
        HbA1C, C-peptide, Red Blood Cell Count, Hematocrit,\\
        Hemoglobin, MCH, MCHC, MCV, MPV, Platelet Count,\\
        Monocytes, Neutrophils, INR, Fibrinogen, Ferritin,\\
        Transferrin, Creatine Kinase
        \end{tabular} \\ \bottomrule
    \end{tabular}
    \end{adjustbox}
\end{table}

The resulting multi-center cohort contains 201,322 patients (median age: 74, female: male = 90,317: 111,005) from four geographic sites of Mayo Clinic, including Rochester: n = 93,558 (46.5\%), MCHS: n = 57,457 (28.5\%), Arizona: n = 25,418 (12.6\%), and Florida: n = 24,889 (12.4\%). For each cancer risk prediction task, data from the same non-cancer group of 177,081 patients (median age: 75, female: male = 79,003: 98,078) were used as the control group. The case groups comprise 6,017 pancreatic cancer patients (prevalence: 3.0\%), 5,330 liver cancer patients (prevalence: 2.6\%), and 13,043 intestinal cancer patients (prevalence: 6.5\%), respectively, of whom 149 patients were diagnosed with multiple cancers and were therefore counted in corresponding binary classification tasks. The resulting EHR dataset includes $2698$ unique 3-character diagnosis codes and $31$ laboratory tests, representing a vocabulary size of $2729$.

\subsection{Baselines}
We implemented three established graph neural networks (GraphSAGE~\cite{hamilton2017inductive}, GAT~\cite{velivckovic2017graph} and GCN~\cite{kipf2016semi}), as well as two sequence models (Transformer and LSTM), as comparison baselines. The baseline GNN models used the previously constructed patient-level EHR graphs as input, without the context-aware, time-decay message passing. Meanwhile, the bucketized heterogeneous EHR encoding matrix served as input to the two sequence baseline models. For all GNN and Transformer models, we used the same multi-head attention-based feature pooling to aggregate patient-level features, while for LSTM, the last hidden state was used as the patient-level representation for classification.

\subsection{Model Training and Evaluation}
We employed a geographic internal-external validation strategy to train and evaluate model generalization, namely leave-one-site-out (LOSO) testing. In each LOSO iteration, data from care sites Mayo Clinic Health System (MCHS), Arizona, and Florida were respectively held out in turn as an independent \textit{test set}, comprising three geographic holdout experiments. Data from the Rochester site, which contributed the largest patient population, were retained for model training in all LOSO iterations, together with the two non-held-out sites, forming the \textit{development set}. This design allowed most available data to be used for model development while iteratively evaluating model generalization to a geographically and operationally distinct healthcare center. Using LOSO testing, we evaluated the model's performance with 12-, 24-, and 36-month diagnostic lead times.

During model training, we also applied 10-fold cross-validation (CV) to the development sets. To mitigate data imbalance, we randomly downsampled the control group during training in each CV-fold. Following the same data exclusion strategy in~\cite{placido2023deep, zheng2025pancreatic, varghese2026digitally}, we used all clinical history up to the index date to train the model (excluding data on the index date and onward). Then, the trained models were tested on data with the diagnostic lead time applied. This strategy has been shown to enhance performance by exposing the model to near-diagnosis feature patterns in training, while maintaining testing legitimacy and objectivity~\cite {placido2023deep, zheng2025pancreatic}. We adopted the AdamW optimizer~\cite{loshchilov2019decoupled} to minimize the binary cross-entropy loss for 12 epochs. The learning rate was reduced by a factor of 10 when the validation loss plateaued for 2 epochs. The GELU activation was used in the final classification modules. 

Hyperparameters for GNN models were determined based on validation performance. Candidate parameters (with the searched sets in parentheses) included time-bucket duration days (\{7, 30, 90\}); node-feature and graph embedding dimensions (\{32, 64, 128, 256\}); message-passing depth and feature pooling attention heads (\{1, 2, 3, 4\}); initial learning rate (\{1e-4, 5e-4, 8e-4, 1e-3, 2e-3, 5e-3\}); mini-batch size (\{8, 16, 32, 64, 128\}); and the training downsampling ratio (\{1:1, 1:2, 1:3, 1:5, 1:10, original\}), where original denotes training without downsampling. The Transformer model used hyperparameters previously optimized in~\cite{varghese2026digitally}, while the LSTM model was separately tuned based on validation performance.

\begin{figure*}[!t]
    \centering
    \captionsetup[subfloat]{position=top}
    \subfloat[Pancreatic cancer]{%
        \includegraphics[width=0.325\textwidth]{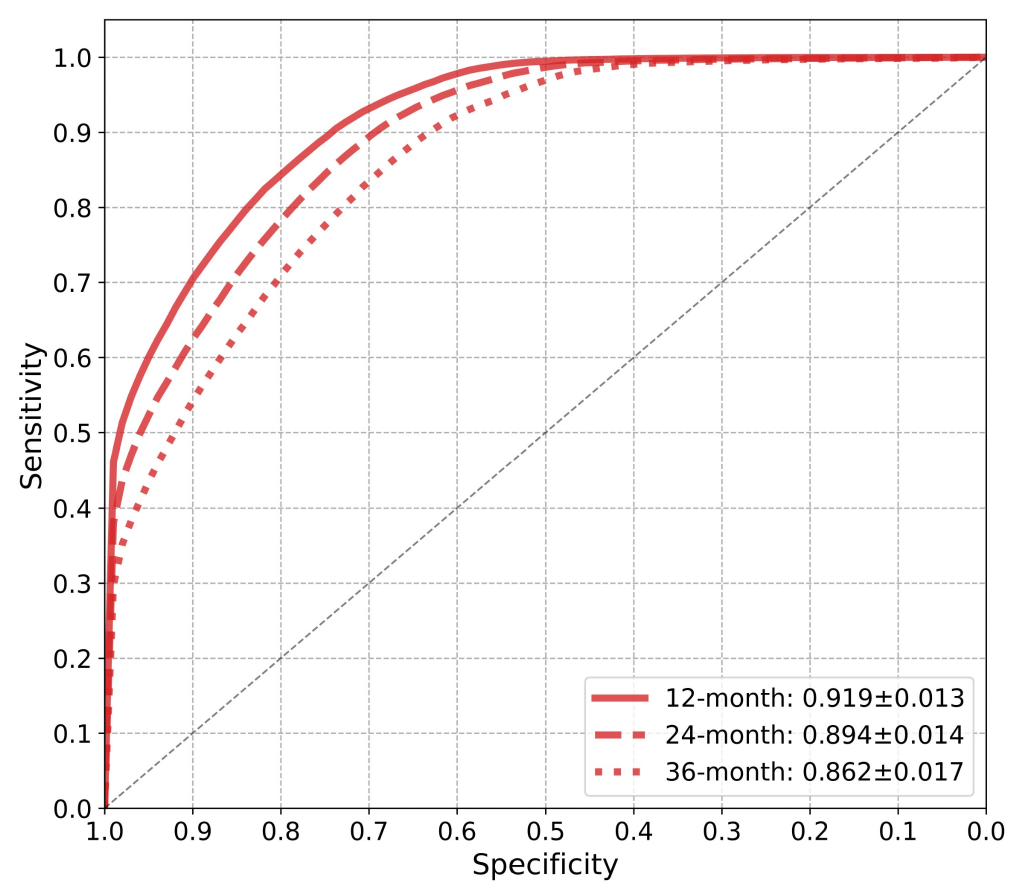}
    }
    \hfill
    \subfloat[Liver cancer]{%
        \includegraphics[width=0.325\textwidth]{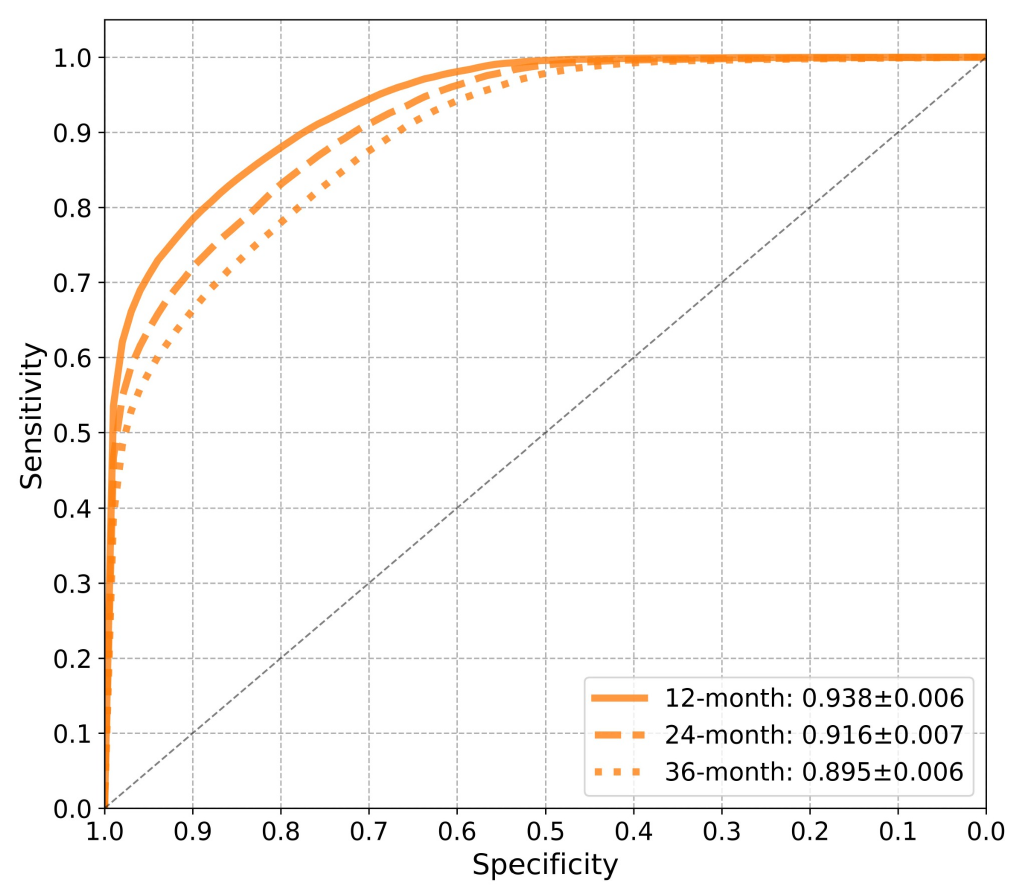}
    }
    \hfill
    \subfloat[Intestinal cancer]{%
        \includegraphics[width=0.325\textwidth]{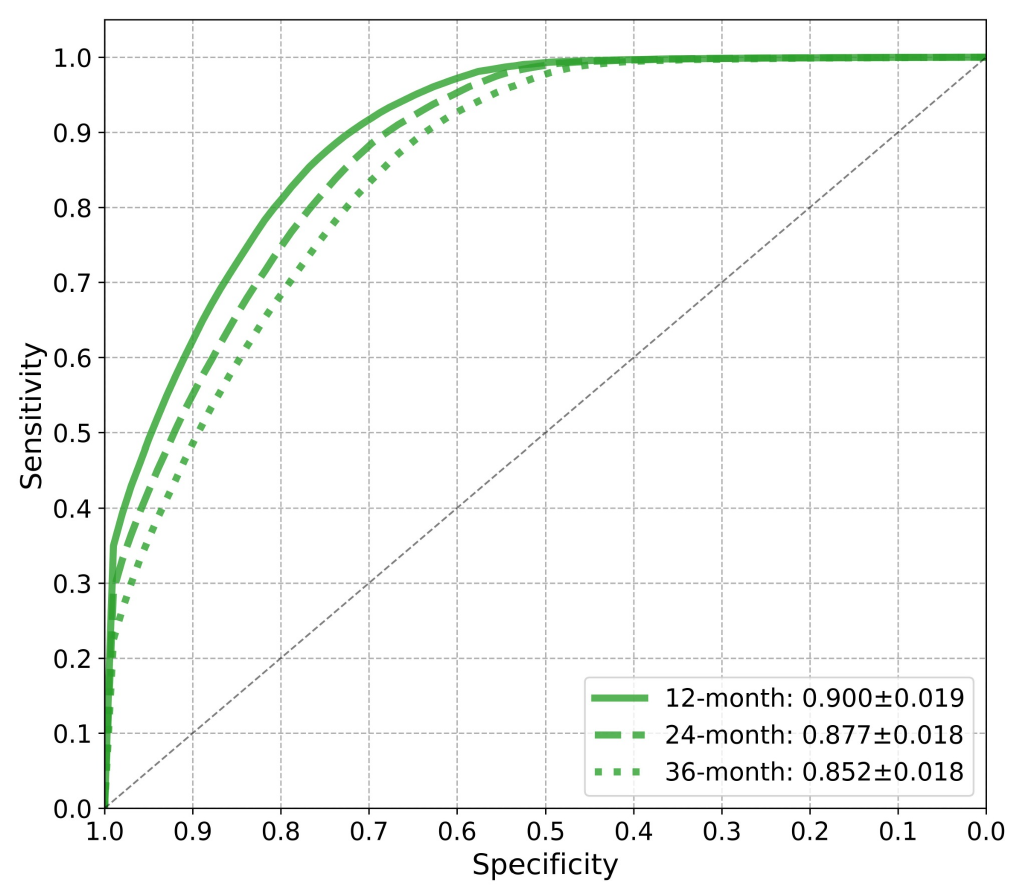}
    }
    
    \subfloat{%
        \includegraphics[width=0.325\textwidth]{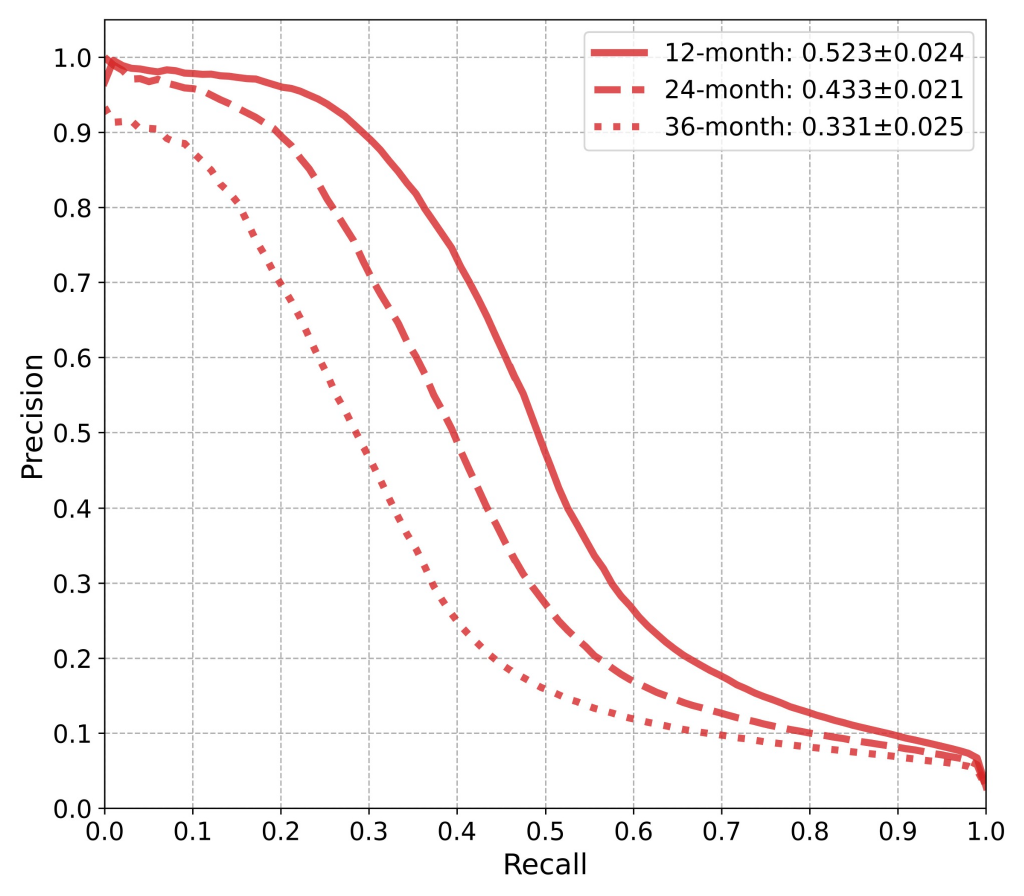}
    }
    \hfill
    \subfloat{%
        \includegraphics[width=0.325\textwidth]{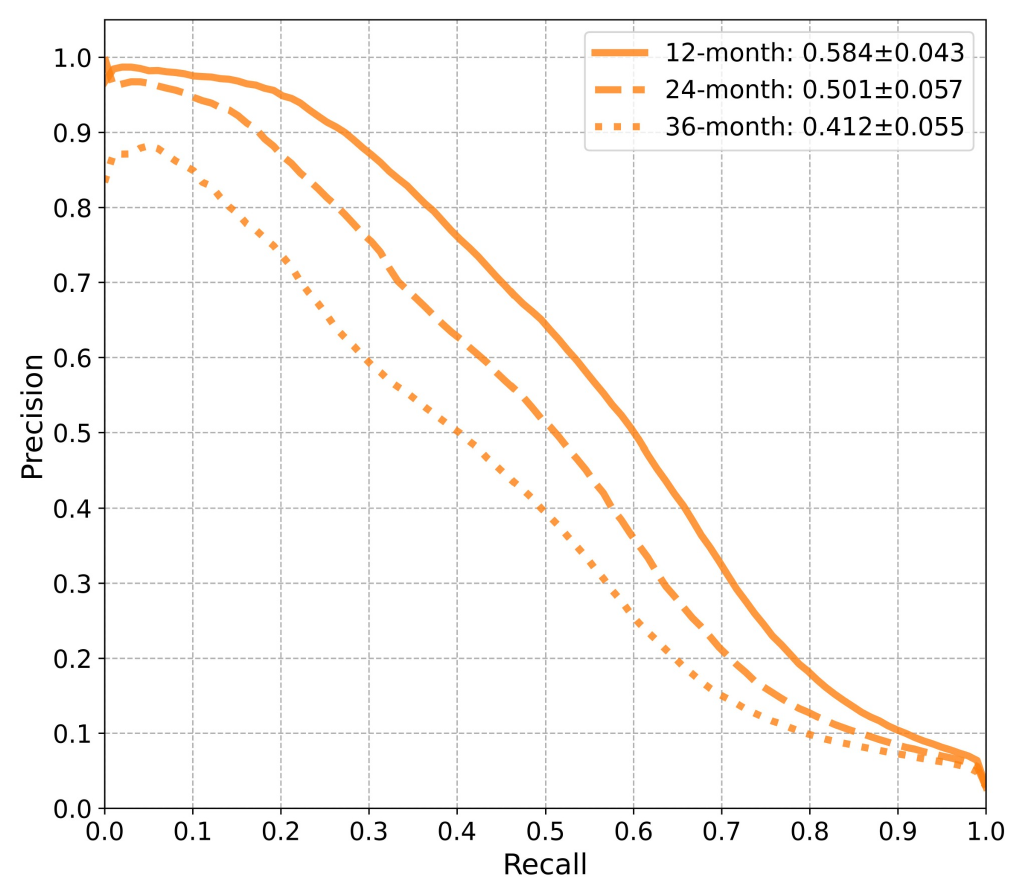}
    }
    \hfill
    \subfloat{%
        \includegraphics[width=0.325\textwidth]{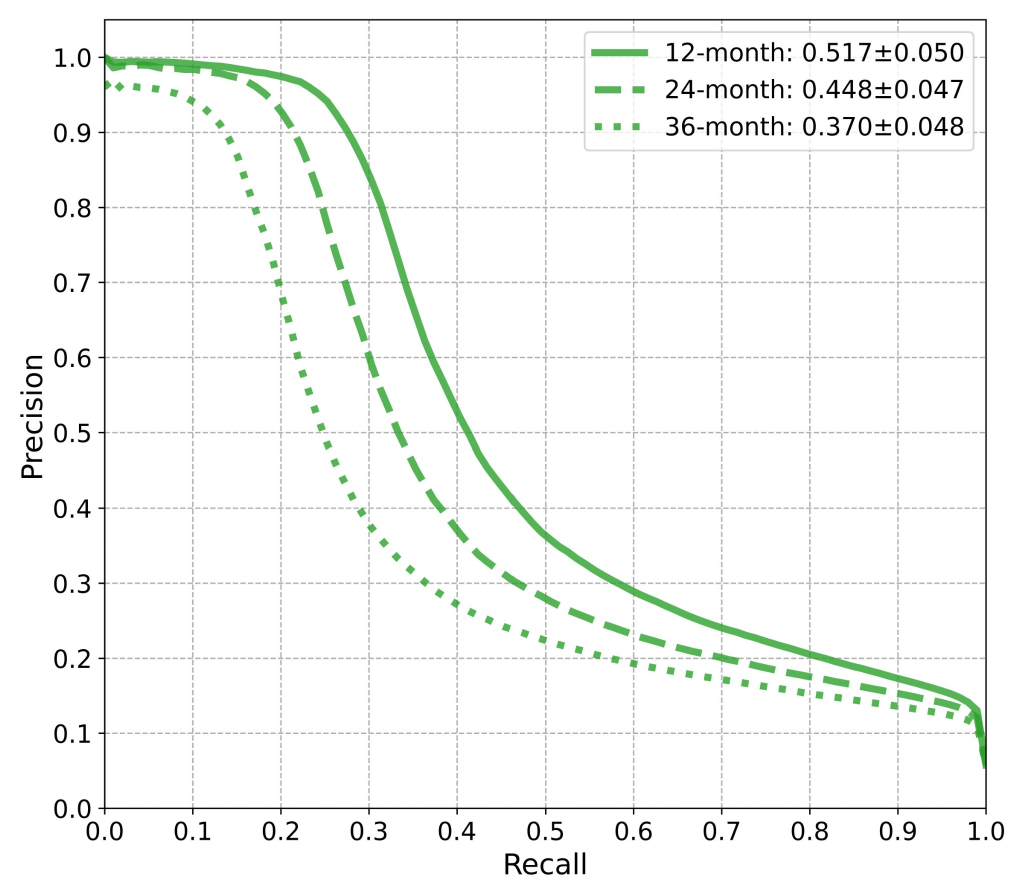}
    }
    \caption{Mean ($\pm$ SD) leave-one-site-out receiver operating characteristic curves (top row) and precision--recall curves (bottom row) for GraphBEAT at 12-, 24-, and 36-month prediction lead times. Columns correspond to (a) pancreatic, (b) liver, and (c) intestinal cancers,
    respectively.}
    \label{fig2_roc_prc}
\end{figure*}

The 10-fold CV produced 10 trained models per LOSO iteration per diagnostic lead time. They were respectively evaluated on the corresponding test sets using the original data distributions and excluding all data within the specified lead-time interval before the index date. In this work, we specifically selected the area under the receiver operating characteristic curve (AUROC) and average precision (AP) as the primary performance metrics because they are invariant to the chosen decision threshold, therefore providing a threshold-independent performance evaluation. The number of floating-point operations (FLOPs) and learnable parameters were used to quantify the computational complexity. The mean and 95\% confidence interval (CI) of metrics obtained from the 10 trained models were recorded and reported. The model explainability was examined using PyTorch Geometric Explainer~\cite{Fey2019Fast} with the GNNExplainer algorithm~\cite{ying2019gnnexplainer}. For each node and edge in a graph, the explainer assigns a score between 0 and 1 to represent its relative importance in the particular graph-level classification. Those scores were then aggregated across graphs to create overall feature importance.

All models were implemented with PyTorch v2.8.0 and PyTorch Geometric v2.8.0.post1 (for GNNs). All experiments were performed on a single NVIDIA H100 80G GPU.

\begin{table*}[]
\centering
\caption{Performance comparison of GraphBEAT and baseline models for pancreatic, liver, and intestinal cancer risk prediction across different diagnostic lead times. } 
\label{tab:results}
\begin{adjustbox}{max width=1\textwidth,center}
\begin{tabular}{@{}ll
                c@{\hspace{0.8em}}c
                @{\hspace{2.0em}}
                c@{\hspace{0.8em}}c
                @{\hspace{2.0em}}
                c@{\hspace{0.8em}}c@{}}
\toprule
Cancer & Model
& \multicolumn{2}{c@{\hspace{2.0em}}} {Lead Time: 12-month}
& \multicolumn{2}{@{}c@{\hspace{2.0em}}} {Lead Time: 24-month}
& \multicolumn{2}{@{}c@{}} {Lead Time: 36-month} \\

\cmidrule(lr){3-4}
\cmidrule(lr){5-6}
\cmidrule(lr){7-8}
& & AUROC & AP & AUROC & AP & AUROC & AP \\
\midrule
    \begin{tabular}[c]{@{}l@{}}\textbf{Pancreas}\\ (6017:177081)\end{tabular} & \begin{tabular}[c]{@{}l@{}} \textbf{GraphBEAT}\\~\end{tabular} & 
    \begin{tabular}[c]{@{}l@{}}\textbf{0.919}\\ {[}0.914, 0.924{]}\end{tabular} & 
    \begin{tabular}[c]{@{}l@{}}\textbf{0.523}\\ {[}0.514, 0.532{]}\end{tabular} & 
    \begin{tabular}[c]{@{}l@{}}\textbf{0.894}\\ {[}0.889, 0.899{]}\end{tabular} & 
    \begin{tabular}[c]{@{}l@{}}\textbf{0.433}\\ {[}0.425, 0.441{]}\end{tabular} & 
    \begin{tabular}[c]{@{}l@{}}\textbf{0.862}\\ {[}0.856, 0.869{]}\end{tabular} & 
    \begin{tabular}[c]{@{}l@{}}\textbf{0.331}\\ {[}0.321, 0.340{]}\end{tabular}    \\
    & \begin{tabular}[c]{@{}l@{}} GraphSAGE\\~\end{tabular}    
    & \begin{tabular}[c]{@{}l@{}}0.913\\ {[}0.907, 0.919{]}\end{tabular}    & 
    \begin{tabular}[c]{@{}l@{}}0.499\\ {[}0.489, 0.509{]}\end{tabular}    & 
    \begin{tabular}[c]{@{}l@{}}0.886\\ {[}0.879, 0.893{]}\end{tabular}    & 
    \begin{tabular}[c]{@{}l@{}}0.399\\ {[}0.387, 0.411{]}\end{tabular}    & 
    \begin{tabular}[c]{@{}l@{}}0.855\\ {[}0.847, 0.863{]}\end{tabular}    & 
    \begin{tabular}[c]{@{}l@{}}0.304\\ {[}0.295, 0.313{]}\end{tabular}    \\
    & \begin{tabular}[c]{@{}l@{}} GAT\\~\end{tabular}  & 
    \begin{tabular}[c]{@{}l@{}}0.908\\ {[}0.901, 0.915{]}\end{tabular}    & 
    \begin{tabular}[c]{@{}l@{}}0.493\\ {[}0.484, 0.501{]}\end{tabular}    & 
    \begin{tabular}[c]{@{}l@{}}0.880\\ {[}0.872, 0.888{]}\end{tabular}    & 
    \begin{tabular}[c]{@{}l@{}}0.393\\ {[}0.381, 0.404{]}\end{tabular}    & 
    \begin{tabular}[c]{@{}l@{}}0.846\\ {[}0.837, 0.854{]}\end{tabular}    & 
    \begin{tabular}[c]{@{}l@{}}0.292\\ {[}0.281, 0.304{]}\end{tabular}  \\
    & \begin{tabular}[c]{@{}l@{}} GCN\\~\end{tabular}  & 
    \begin{tabular}[c]{@{}l@{}}0.901\\ {[}0.894, 0.907{]}\end{tabular}    & 
    \begin{tabular}[c]{@{}l@{}}0.460\\ {[}0.453, 0.467{]}\end{tabular}    & 
    \begin{tabular}[c]{@{}l@{}}0.874\\ {[}0.867, 0.881{]}\end{tabular}    & 
    \begin{tabular}[c]{@{}l@{}}0.373\\ {[}0.366, 0.381{]}\end{tabular}    & 
    \begin{tabular}[c]{@{}l@{}}0.841\\ {[}0.832, 0.850{]}\end{tabular}    & 
    \begin{tabular}[c]{@{}l@{}}0.285\\ {[}0.277, 0.294{]}\end{tabular}    \\
    & \begin{tabular}[c]{@{}l@{}} Transformer\\~\end{tabular}  & 
    \begin{tabular}[c]{@{}l@{}}0.837\\ {[}0.827, 0.848{]}\end{tabular} & 
    \begin{tabular}[c]{@{}l@{}}0.341\\ {[}0.330, 0.353{]}\end{tabular} & 
    \begin{tabular}[c]{@{}l@{}}0.797\\ {[}0.782, 0.813{]}\end{tabular} & 
    \begin{tabular}[c]{@{}l@{}}0.260\\ {[}0.247, 0.273{]}\end{tabular} & 
    \begin{tabular}[c]{@{}l@{}}0.760\\ {[}0.745, 0.776{]}\end{tabular} & 
    \begin{tabular}[c]{@{}l@{}}0.206\\ {[}0.196, 0.216{]}\end{tabular} \\
    & \begin{tabular}[c]{@{}l@{}} LSTM\\~\end{tabular}  & 
    \begin{tabular}[c]{@{}l@{}}0.808\\ {[}0.788, 0.827{]}\end{tabular}    & 
    \begin{tabular}[c]{@{}l@{}}0.264\\ {[}0.242, 0.285{]}\end{tabular}    & 
    \begin{tabular}[c]{@{}l@{}}0.777\\ {[}0.757, 0.797{]}\end{tabular}    & 
    \begin{tabular}[c]{@{}l@{}}0.212\\ {[}0.195, 0.228{]}\end{tabular}    & 
    \begin{tabular}[c]{@{}l@{}}0.745\\ {[}0.725, 0.766{]}\end{tabular}    & 
    \begin{tabular}[c]{@{}l@{}}0.167\\ {[}0.153, 0.182{]}\end{tabular}    \\
    
    \addlinespace[4pt]
    \begin{tabular}[c]{@{}l@{}}\textbf{Liver}\\(5330:177081)\end{tabular}       & 
    \begin{tabular}[c]{@{}l@{}} \textbf{GraphBEAT}\\~\end{tabular} & 
    \begin{tabular}[c]{@{}l@{}}\textbf{0.938}\\ {[}0.935, 0.940{]}\end{tabular} & 
    \begin{tabular}[c]{@{}l@{}}\textbf{0.584}\\ {[}0.567, 0.600{]}\end{tabular}    & 
    \begin{tabular}[c]{@{}l@{}}\textbf{0.916}\\ {[}0.914, 0.919{]}\end{tabular}    & 
    \begin{tabular}[c]{@{}l@{}}\textbf{0.501}\\ {[}0.480, 0.523{]}\end{tabular} & 
    \begin{tabular}[c]{@{}l@{}}\textbf{0.895}\\ {[}0.892, 0.897{]}\end{tabular} & 
    \begin{tabular}[c]{@{}l@{}}\textbf{0.412}\\ {[}0.391, 0.433{]}\end{tabular} \\ & 
    \begin{tabular}[c]{@{}l@{}} GraphSAGE\\~\end{tabular}   & 
    \begin{tabular}[c]{@{}l@{}}0.931\\ {[}0.928, 0.934{]}\end{tabular}    & 
    \begin{tabular}[c]{@{}l@{}}0.543\\ {[}0.526, 0.560{]}\end{tabular}    & 
    \begin{tabular}[c]{@{}l@{}}0.912\\ {[}0.908, 0.916{]}\end{tabular}    & 
    \begin{tabular}[c]{@{}l@{}}0.459\\ {[}0.438, 0.480{]}\end{tabular}    & 
    \begin{tabular}[c]{@{}l@{}}0.889\\ {[}0.887, 0.892{]}\end{tabular}    & 
    \begin{tabular}[c]{@{}l@{}}0.370\\ {[}0.351, 0.389{]}\end{tabular}    \\
    & \begin{tabular}[c]{@{}l@{}} GAT\\~\end{tabular}  & 
    \begin{tabular}[c]{@{}l@{}}0.927\\ {[}0.924, 0.930{]}\end{tabular}    & 
    \begin{tabular}[c]{@{}l@{}}0.534\\ {[}0.513, 0.556{]}\end{tabular}    & 
    \begin{tabular}[c]{@{}l@{}}0.905\\ {[}0.901, 0.908{]}\end{tabular}    & 
    \begin{tabular}[c]{@{}l@{}}0.443\\ {[}0.419, 0.467{]}\end{tabular}    & 
    \begin{tabular}[c]{@{}l@{}}0.884\\ {[}0.881, 0.888{]}\end{tabular}    & 
    \begin{tabular}[c]{@{}l@{}}0.368\\ {[}0.347, 0.389{]}\end{tabular}    \\
    & \begin{tabular}[c]{@{}l@{}} GCN\\~\end{tabular} & 
    \begin{tabular}[c]{@{}l@{}}0.923\\ {[}0.920, 0.926{]}\end{tabular}    & 
    \begin{tabular}[c]{@{}l@{}}0.521\\ {[}0.509, 0.533{]}\end{tabular}    & 
    \begin{tabular}[c]{@{}l@{}}0.902\\ {[}0.899, 0.905{]}\end{tabular}    & 
    \begin{tabular}[c]{@{}l@{}}0.450\\ {[}0.438, 0.462{]}\end{tabular}    & 
    \begin{tabular}[c]{@{}l@{}}0.881\\ {[}0.879, 0.883{]}\end{tabular}    & 
    \begin{tabular}[c]{@{}l@{}}0.368\\ {[}0.354, 0.382{]}\end{tabular}    \\
    & \begin{tabular}[c]{@{}l@{}} Transformer\\~\end{tabular}  & 
    \begin{tabular}[c]{@{}l@{}}0.853\\ {[}0.846, 0.860{]}\end{tabular}    & 
    \begin{tabular}[c]{@{}l@{}}0.382\\ {[}0.356, 0.407{]}\end{tabular}    & 
    \begin{tabular}[c]{@{}l@{}}0.830\\ {[}0.821, 0.840{]}\end{tabular}    & 
    \begin{tabular}[c]{@{}l@{}}0.324\\ {[}0.297, 0.350{]}\end{tabular}    & 
    \begin{tabular}[c]{@{}l@{}}0.806\\ {[}0.793, 0.818{]}\end{tabular}    & 
    \begin{tabular}[c]{@{}l@{}}0.275\\ {[}0.247, 0.303{]}\end{tabular}    \\
    & \begin{tabular}[c]{@{}l@{}} LSTM\\~\end{tabular}         & 
    \begin{tabular}[c]{@{}l@{}}0.849\\ {[}0.846, 0.852{]}\end{tabular}    &
    \begin{tabular}[c]{@{}l@{}}0.347\\ {[}0.326, 0.368{]}\end{tabular}    & 
    \begin{tabular}[c]{@{}l@{}}0.834\\ {[}0.832, 0.837{]}\end{tabular}    & 
    \begin{tabular}[c]{@{}l@{}}0.292\\ {[}0.271, 0.313{]}\end{tabular}    & 
    \begin{tabular}[c]{@{}l@{}}0.808\\ {[}0.806, 0.810{]}\end{tabular}    & 
    \begin{tabular}[c]{@{}l@{}}0.248  \\ {[}0.229, 0.268{]}\end{tabular}  \\
    
    \addlinespace[4pt]
    \begin{tabular}[c]{@{}l@{}}\textbf{Intestine}\\ (13043:177081)\end{tabular} & 
    \begin{tabular}[c]{@{}l@{}} \textbf{GraphBEAT}\\~\end{tabular} &
    \begin{tabular}[c]{@{}l@{}}\textbf{0.900}\\ {[}0.893, 0.907{]}\end{tabular}  & 
    \begin{tabular}[c]{@{}l@{}}\textbf{0.517}\\ {[}0.498, 0.536{]}\end{tabular} & 
    \begin{tabular}[c]{@{}l@{}}\textbf{0.877}\\ {[}0.870, 0.884{]}\end{tabular} & 
    \begin{tabular}[c]{@{}l@{}}\textbf{0.448}\\ {[}0.430, 0.466{]}\end{tabular}    & 
    \begin{tabular}[c]{@{}l@{}}\textbf{0.852}\\ {[}0.845, 0.859{]}\end{tabular} & 
    \begin{tabular}[c]{@{}l@{}}\textbf{0.370}\\ {[}0.351, 0.388{]}\end{tabular}    \\& 
    \begin{tabular}[c]{@{}l@{}} GraphSAGE\\~\end{tabular}  & 
    \begin{tabular}[c]{@{}l@{}}0.894\\ {[}0.886, 0.901{]}\end{tabular}    & 
    \begin{tabular}[c]{@{}l@{}}0.489\\ {[}0.471, 0.507{]}\end{tabular}    & 
    \begin{tabular}[c]{@{}l@{}}0.871\\ {[}0.863, 0.879{]}\end{tabular}    & 
    \begin{tabular}[c]{@{}l@{}}0.417\\ {[}0.397, 0.437{]}\end{tabular}    & 
    \begin{tabular}[c]{@{}l@{}}0.843\\ {[}0.835, 0.852{]}\end{tabular}    & 
    \begin{tabular}[c]{@{}l@{}}0.334\\ {[}0.315, 0.353{]}\end{tabular}    \\
    & \begin{tabular}[c]{@{}l@{}} GAT\\~\end{tabular}          & 
    \begin{tabular}[c]{@{}l@{}}0.887\\ {[}0.878, 0.895{]}\end{tabular}    & 
    \begin{tabular}[c]{@{}l@{}}0.469\\ {[}0.451, 0.487{]}\end{tabular}    & 
    \begin{tabular}[c]{@{}l@{}}0.863\\ {[}0.853, 0.872{]}\end{tabular}    & 
    \begin{tabular}[c]{@{}l@{}}0.400\\ {[}0.379, 0.421{]}\end{tabular}    & 
    \begin{tabular}[c]{@{}l@{}}0.833\\ {[}0.823, 0.843{]}\end{tabular}    & 
    \begin{tabular}[c]{@{}l@{}}0.317\\ {[}0.298, 0.335{]}\end{tabular}    \\
    & \begin{tabular}[c]{@{}l@{}} GCN\\~\end{tabular}          & 
    \begin{tabular}[c]{@{}l@{}}0.881\\ {[}0.873, 0.890{]}\end{tabular}    & 
    \begin{tabular}[c]{@{}l@{}}0.451\\ {[}0.426, 0.476{]}\end{tabular}    & 
    \begin{tabular}[c]{@{}l@{}}0.859\\ {[}0.850, 0.867{]}\end{tabular}    & 
    \begin{tabular}[c]{@{}l@{}}0.380\\ {[}0.356, 0.404{]}\end{tabular}    & 
    \begin{tabular}[c]{@{}l@{}}0.831\\ {[}0.822, 0.840{]}\end{tabular}    & 
    \begin{tabular}[c]{@{}l@{}}0.304\\ {[}0.281, 0.327{]}\end{tabular}    \\
    & \begin{tabular}[c]{@{}l@{}} Transformer\\~\end{tabular}  & 
    \begin{tabular}[c]{@{}l@{}}0.760\\ {[}0.745, 0.776{]}\end{tabular}    & 
    \begin{tabular}[c]{@{}l@{}}0.258\\ {[}0.239, 0.277{]}\end{tabular}    & 
    \begin{tabular}[c]{@{}l@{}}0.734\\ {[}0.718, 0.750{]}\end{tabular}    & 
    \begin{tabular}[c]{@{}l@{}}0.201\\ {[}0.184, 0.217{]}\end{tabular}    & 
    \begin{tabular}[c]{@{}l@{}}0.706\\ {[}0.692, 0.719{]}\end{tabular}    & 
    \begin{tabular}[c]{@{}l@{}}0.154\\ {[}0.142, 0.166{]}\end{tabular}    \\
    & \begin{tabular}[c]{@{}l@{}} LSTM\\~\end{tabular}         & 
    \begin{tabular}[c]{@{}l@{}}0.830\\ {[}0.813, 0.848{]}\end{tabular}    & 
    \begin{tabular}[c]{@{}l@{}}0.358\\ {[}0.325, 0.390{]}\end{tabular}    & 
    \begin{tabular}[c]{@{}l@{}}0.801\\ {[}0.782, 0.820{]}\end{tabular}    & 
    \begin{tabular}[c]{@{}l@{}}0.290\\ {[}0.259, 0.320{]}\end{tabular}    & 
    \begin{tabular}[c]{@{}l@{}}0.761\\ {[}0.739, 0.783{]}\end{tabular}    & 
    \begin{tabular}[c]{@{}l@{}}0.225\\ {[}0.197, 0.253{]}\end{tabular}    \\ 
    \cmidrule(r){1-8}
\end{tabular}
\end{adjustbox}
\vspace{2pt}
\begin{minipage}{0.85\textwidth}
\footnotesize
\textit{Note:} Values are reported as the mean [95\% confidence interval] across the three leave-one-site-out test sets; Case-control numbers are shown below the cancer type. AUROC: area under the receiver operating characteristic curve; AP: average precision.
\end{minipage}
\end{table*}

\subsection{Performance Results}
Figure~\ref{fig2_roc_prc} presents the mean receiver operating characteristic curves and precision-recall curves of the GraphBEAT model across the 3 held-out test sites for pancreatic, liver, and intestinal cancer risk prediction, with 12-, 24-, and 36-month lead times, respectively. Consistent with observations in previous studies~\cite{placido2023deep, zheng2025pancreatic, varghese2026digitally}, predictive performance decreased with longer lead times as the task became more challenging, but GraphBEAT maintained strong discrimination across all outcomes. At the 12-, 24-, and 36-month diagnostic lead times, respectively, GraphBEAT achieved the mean LOSO-AUROC of $0.919$, $0.894$, and $0.862$ for pancreatic cancer; $0.938$, $0.916$, and $0.895$ for liver cancer; and $0.900$, $0.877$, and $0.852$ for intestinal cancer. The corresponding mean LOSO-AP values were $0.523$, $0.433$, and $0.331$ for pancreatic cancer; $0.584$, $0.501$, and $0.412$ for liver cancer; and $0.517$, $0.448$, and $0.370$ for intestinal cancer. Despite having the lowest prevalence in our dataset, liver cancer prediction yielded the highest AUROC and AP at every lead time.

The LOSO testing results also demonstrated consistent performance across geographically distinct test sites. For instance, in the 12-month pancreatic cancer prediction task, GraphBEAT achieved the mean [$95\%$ confidence interval] LOSO-AUROC and LOSO-AP of $0.936~[0.934-0.937]$ and $0.536~[0.524-0.547]$ on the MCHS test set, $0.916~[0.913-0.919]$ and $0.522~[0.512-0.533]$ on the Arizona set, and $0.906~[0.901-0.910]$ and $0.511~[0.488-0.530]$ on the Florida set. The higher discrimination observed at MCHS may suggest potential relevance to community-based healthcare patterns, which could be more clinically valuable for model utility. The same site-wise performance pattern can also be observed for liver and intestinal cancers, which supports the geographic generalizability of GraphBEAT.

Table~\ref{tab:results} further compares the LOSO testing performance of GraphBEAT with the GraphSAGE, GAT, GCN, Transformer, and LSTM baselines. Overall, GraphBEAT outperformed all GNN and sequence models in all subtasks. Compared with the strongest baseline GNN model, GraphSAGE, at each task and diagnostic lead time, GraphBEAT improved the absolute cross-site AP by $0.024$ to $0.042$ and AUROC by $0.004$ to $0.009$. The absolute improvements from the strongest sequence baseline were more pronounced, ranging from $0.125$ to $0.202$ in AP and from $0.070$ to $0.102$ in AUROC. It should be noted that AP is more relevant for evaluating risk stratification performance in rare diseases, as it complements AUROC by directly reflecting the precision-recall trade-off, which focuses on the minority class. At the 12-month horizon, GraphBEAT achieved LOSO-AP of $0.523$, $0.584$, and $0.517$ for pancreatic, liver, and intestinal cancers, respectively, improving by 4.8\% to 7.6\% compared with baseline models across all experimental setups. Additionally, with the proposed patient EHR graph structure, all GNN models demonstrated superior performance to the sequence models by a material margin, while GraphBEAT provided a consistent, additional improvement over the established GNN architectures.

\begin{figure}[!ht]
    \centering
    \includegraphics[width=1\linewidth]{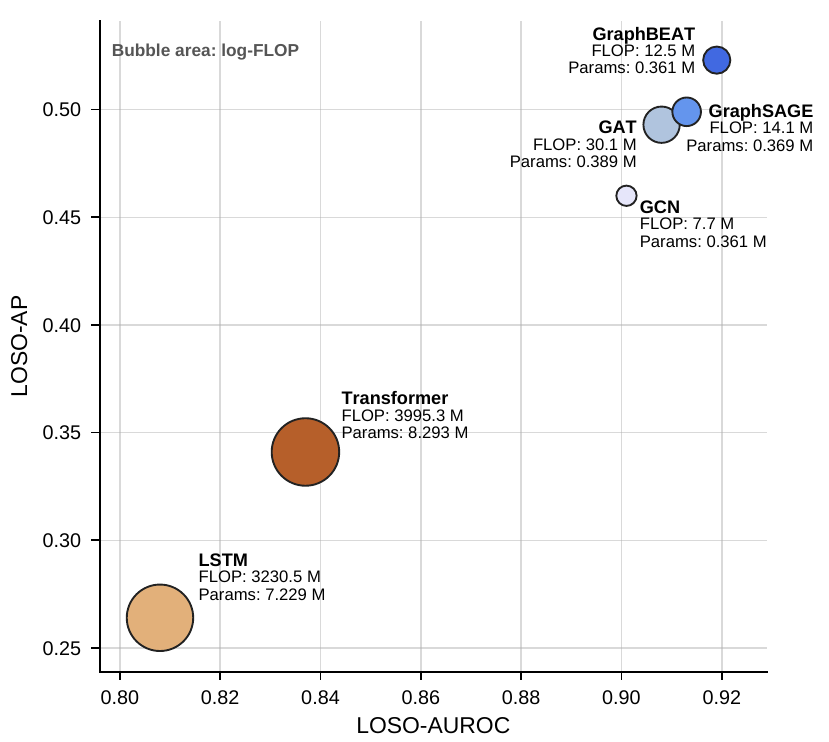}
    \caption{Comparison of computational complexity and predictive performance between GraphBEAT and baseline models for pancreatic cancer risk prediction with a 12-month lead time. The x- and y-axes show the mean leave-one-site-out (LOSO) AUROC and average precision (AP), respectively. Bubble area represents the logarithmically scaled number of floating-point operations (FLOPs). Annotations report the FLOPs and number of learnable parameters in millions (M).}
    \label{fig4}
\end{figure}

\begin{table}[!ht]
    \centering
    \caption{Ablation studies on the MCHS test set for 12-month pancreatic cancer risk prediction.}
    \label{tab:ablation}
    \begin{adjustbox}{max width=1\columnwidth,center}
    \begin{tabular}{@{}lcc@{}}
        \toprule
        Configuration & AUROC & AP \\
        \midrule
        \multicolumn{3}{@{}l}{\textit{Node feature selection}} \\
        \addlinespace[2pt]
        Feature + Value + Time & \textbf{0.936} & \textbf{0.536} \\
        Feature + Value  & 0.922 & 0.499 \\
        Feature + Time  & 0.729 & 0.086 \\
        Value + Time & 0.881 & 0.408 \\
        Feature  & 0.712 & 0.087 \\
        Value  & 0.815 & 0.185 \\
        Time  & 0.549 & 0.032 \\

        \addlinespace[3pt]
        \multicolumn{3}{@{}l}{\textit{Feature embedding selection}} \\
        \addlinespace[2pt]
        \begin{tabular}[t]{@{}l@{}} Shared identity embedding \& \\ 
        Shared value mapping \end{tabular}  & \textbf{0.936} & \textbf{0.536}  \\   
        Separate identity embeddings  & 0.931 & 0.476  \\
        Separate value mappings    & 0.912 & 0.385  \\

        \addlinespace[3pt]
        \multicolumn{3}{@{}l}{\textit{Edge weight formulation}} \\
        \addlinespace[2pt]
        Similarity$_{\text{(src, dest)}}$ $\times$ Value$_{\text{src}}$ & 
        \textbf{0.936} & \textbf{0.536} \\
        Similarity$_{\text{(src, dest)}}$ & 0.931 & 0.506 \\
        Value$_{\text{src}}$ & 0.931 & 0.488 \\
        Without edge weight normalization & 0.904 & 0.370 \\

        \addlinespace[3pt]
        \multicolumn{3}{@{}l}{\textit{Time-decay ablation}} \\
        \addlinespace[2pt]
        With time decay & \textbf{0.936} & \textbf{0.536} \\
        Without time decay & 0.924 & 0.452 \\
        \bottomrule
    \end{tabular}
    \end{adjustbox}
\end{table}

\subsection{Computational Complexity Analysis}
Figure~\ref{fig4} illustrates the discriminative performance vs. computational complexity of all GNN and sequence models in the 12-month pancreatic cancer prediction task. Owing to the compact patient-level EHR graph representation, all GNN models were substantially more computationally efficient than the sequence baselines, which on the other hand operated on the full feature-bucket matrix. 

For 12-month pancreatic cancer prediction, GraphBEAT, GraphSAGE, GAT, and GCN required a mean of $12.5$, $14.1$, $30.1$, and $7.7$ million (M) FLOPs per patient, respectively, under the current model parameter configuration. By comparison, Transformer and LSTM required 3995.3M and 3230.5M FLOPs, respectively, which are around $107$ to $521$ times as many FLOPs as the graph models. Moreover, graph models were also considerably smaller, with only around $0.4$M trainable parameters compared with $8.3$M for the Transformer and $7.2$M for the LSTM, representing over a $94\%$ reduction in model size with the graph-based approaches.

Among GNN models, GCN had the lowest computational cost, while GraphBEAT, the second-most efficient model, achieved the best overall performance-efficiency tradeoff. Compared with the best baseline model (i.e., GraphSAGE), GraphBEAT used $11.2\%$ fewer FLOPs and $2.2\%$ fewer parameters while improving AP by $4.8\%$ in the same task. As such, the efficiency gains indicate that GraphBEAT's context-aware, temporal-decay message passing, coupled with one-hop message aggregation, improved predictive performance while maintaining computation comparable to other GNNs and a much lower computational burden than sequence models.

\begin{table}[!ht]
    \centering
    \caption{Hyperparameter selection based on validation average precision (AP) in 12-month pancreatic cancer prediction. }
    \label{tab:hyperparameters}
    \begin{adjustbox}{max width=1\columnwidth,center}
    \begin{tabular}{@{}ll@{}}
    \toprule
    Hyperparameter & Evaluated values and AP (in parentheses) \\ 
    \midrule
    \begin{tabular}[t]{@{}l@{}} Time-bucket duration\\($\tau$ days) \end{tabular}& 
        7 (0.527); \textbf{30} (0.541); 90 (0.502) \\
        \addlinespace[2pt]
        
        \begin{tabular}[t]{@{}l@{}} Node feature dimension\\ ($d_{\mathrm{node}}$) \end{tabular}& 
        32 (0.511); 64 (0.519); \textbf{128} (0.541); 256 (0.476)\\
        \addlinespace[2pt]
        
        \begin{tabular}[t]{@{}l@{}} Graph embedding dimension\\ ($d_{\mathrm{graph}}$)  \end{tabular}& 
        32 (0.505); \textbf{64} (0.541); 128 (0.468); 256 (0.473) \\
        \addlinespace[2pt]
        
        \begin{tabular}[t]{@{}l@{}} Message-passing depth\\ ($K$) \end{tabular}&
        \textbf{1} (0.541); 2 (0.517); 3 (0.524); 4 (0.515) \\
        \addlinespace[2pt]
        
        \begin{tabular}[t]{@{}l@{}} Feature pooling attention heads\\ ($R$) \end{tabular}& 
        1 (0.505); 2 (0.512); \textbf{3} (0.541); 4 (0.517) \\
        \addlinespace[2pt]
        
        Initial learning rate & 
        \begin{tabular}[t]{@{}l@{}}1e-4 (0.511); 5e-4 (0.524); \textbf{8e-4} (0.541); \\ 
        1e-3 (0.456); 2e-3 (0.411); 5e-3 (0.298) \end{tabular}\\
        \addlinespace[2pt]
        Training mini-batch size &
        \begin{tabular}[t]{@{}l@{}} 8 (0.441); \textbf{16} (0.541); 32 (0.533); \\
        64 (0.517); 128 (0.496) \end{tabular}\\
        \addlinespace[2pt]
        Training down-sampling ratio &
        \begin{tabular}[t]{@{}l@{}}$1{:}1$ (0.487); $1{:}2$ (0.502); \textbf{1:3} (0.541); \\
        $1{:}5$ (0.501); $1{:}10$ (0.448); original (0.184) \end{tabular}\\ 
        \bottomrule
    \end{tabular}
    \end{adjustbox}
\end{table}

\begin{figure*}[!t]
    \centering
    \captionsetup[subfloat]{position=top}
    \subfloat[Pancreatic cancer]{%
        \includegraphics[width=0.325\textwidth]{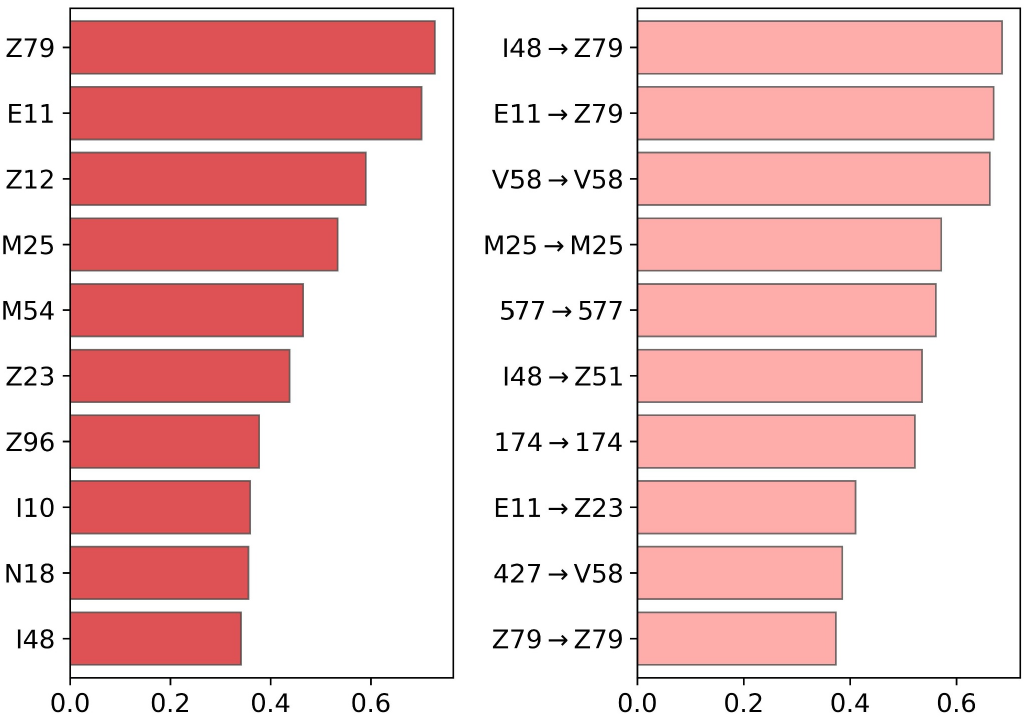}
    }
    \hfill
    \subfloat[Liver cancer]{%
        \includegraphics[width=0.325\textwidth]{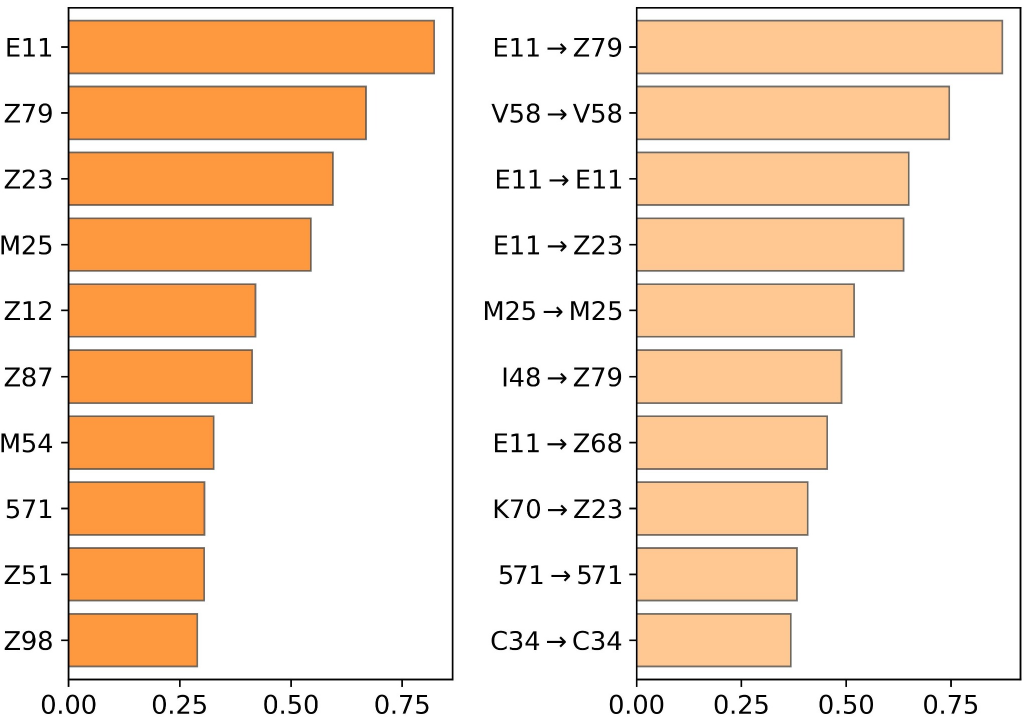}
    }
    \hfill
    \subfloat[Intestinal cancer]{%
        \includegraphics[width=0.325\textwidth]{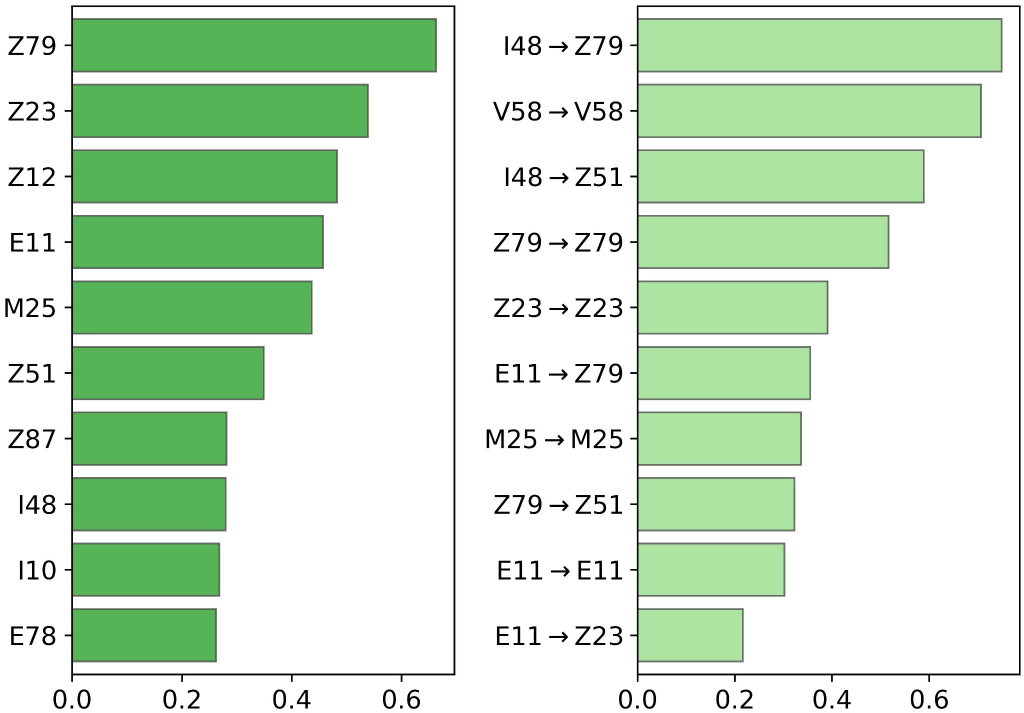}
    }
    \medskip
    
    \subfloat{%
        \includegraphics[width=0.325\textwidth]{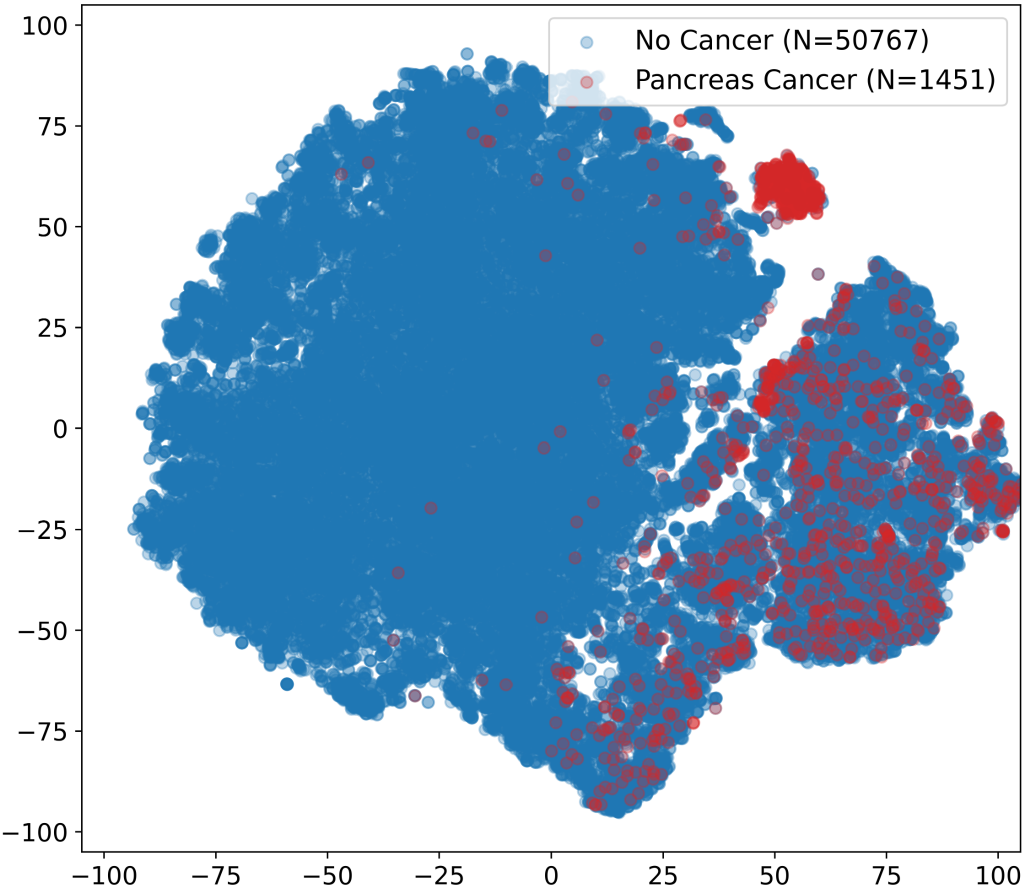}
    }
    \hfill
    \subfloat{%
        \includegraphics[width=0.325\textwidth]{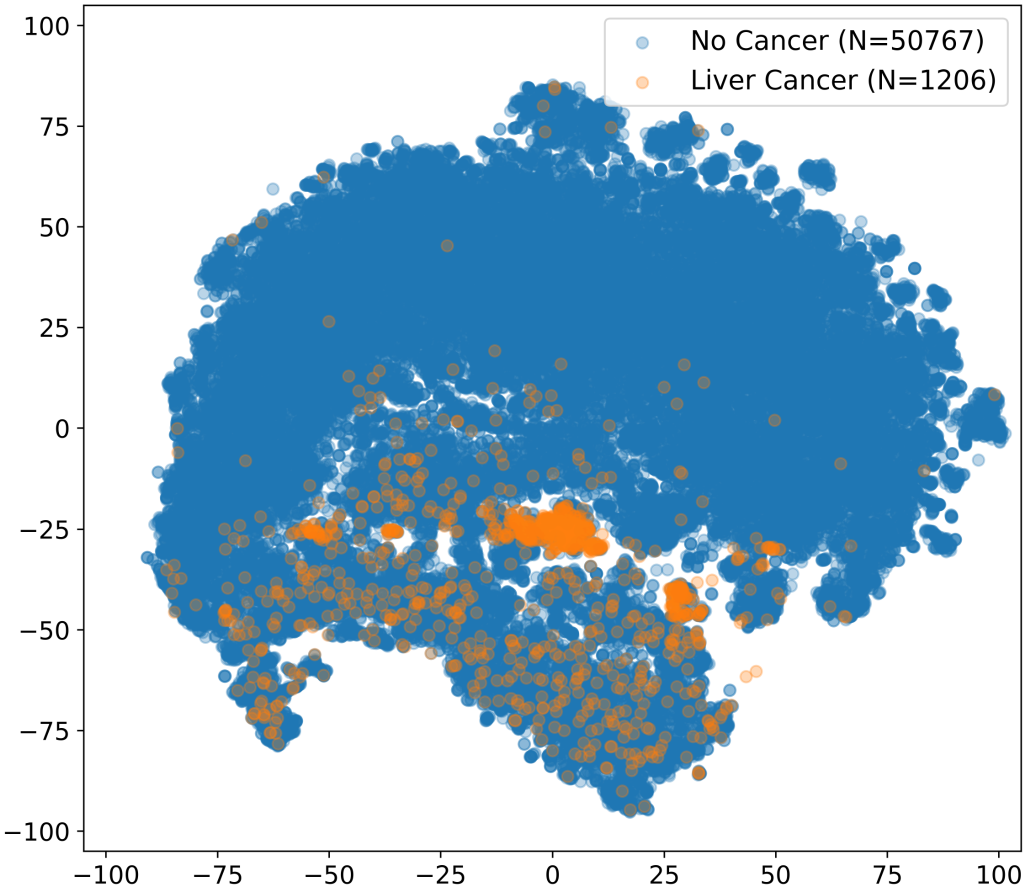}
    }
    \hfill
    \subfloat{%
        \includegraphics[width=0.325\textwidth]{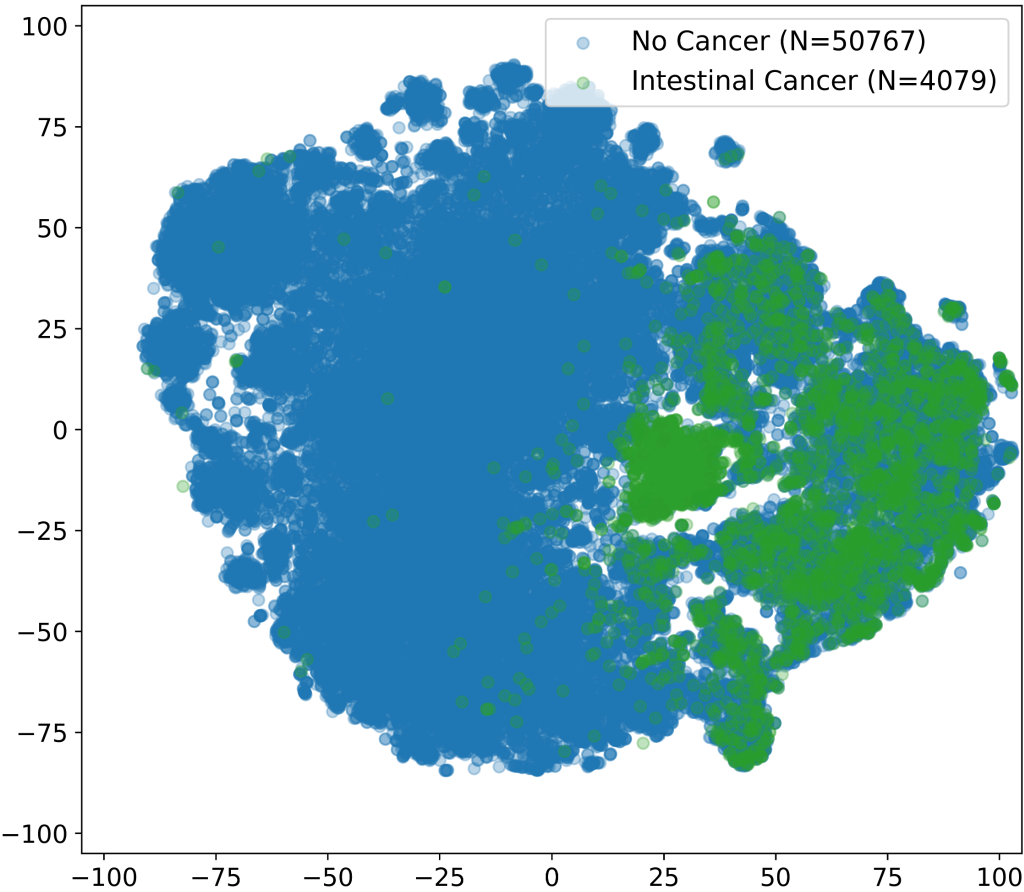}
    }
    \caption{Model explainability analysis for 12-month pancreatic cancer (column a, left), liver cancer (column b, middle), and intestinal cancer (column c, right) risk prediction on the MCHS test sets. Upper row: top-10 contributing nodes and edges for true-positive predictions. Lower row: t-SNE representations of pooled graph embeddings before classification. Each dot represents a patient in the dataset.}
\label{fig5_explain}
\end{figure*}

\subsection{Ablation Studies}
We conducted ablation experiments to investigate three key components of the model design, i.e., node features, edge weights, and time decay. Table~\ref{tab:ablation} summarizes the results obtained from the MCHS test set for pancreatic cancer risk prediction with a 12-month lead time. Note that the MCHS ablations were conducted to characterize the fixed architecture and did not influence model selection. For node feature selection, the combination of feature identity, observed value, and temporal information produced the strongest performance. In contrast, removing any of these components reduced performance, with two-component features generally outperforming the single-component variants. Among the two-component variants, feature identity combined with observed value performed the best (AUROC: $0.922$; AP: $0.499$). Adding temporal information increased AP by $7.4\%$ (from $0.499$ to $0.536$), while using time alone yielded the worst AP (0.032). This suggests that temporal encoding was most informative when integrated with both feature identity and observed values.

Using separate feature-embedding functions for diagnosis and laboratory data, either for feature identity or value mapping, led to suboptimal performance (separate identity embeddings: AP = $0.476$; separate value mappings: AP = $0.385$) compared with sharing these functions across modalities (AP = $0.536$). As GraphBEAT integrates diagnosis and laboratory nodes into a unified graph and uses feature similarity to determine source-target relevance, shared embedding functions may better align diagnosis and blood work in a common latent space and enable direct message passing across heterogeneous data. The notable performance decrease with separate value mappings further suggests that modality-specific transformations may produce incompatible value representations. By contrast, a common value mapping promotes more consistent integration of heterogeneous clinical observations.

Including both contextual similarity and value information in edge weighting yielded the best discriminative performance (AP: $0.536$), improving by $5.9\%$ and $9.8\%$ compared to using only similarity (AP: $0.506$) and value (AP: $0.488$), respectively. By comparison, AUROC differences were at most $0.005$ across edge-weighting strategies, underscoring the importance of referring to precision-recall curve-related metrics in this task. Additionally, removing the weight normalization step in Eq.~\ref{eq:edge_weight} substantially reduced the performance, with the AP dropping by over $30\%$ to $0.370$. 

Lastly, incorporating exponential time decay into message passing resulted in an $18.6\%$ improvement in AP (from $0.452$ to $0.536$), indicating that the decayed influence in message-passing over time is beneficial for the model to extract informative cancer-related features. In summary, the ablation results support the final GraphBEAT design in EHR modeling.

\subsection{Hyperparameter Tuning Results}
Table~\ref{tab:hyperparameters} summarizes the hyperparameter-selection results using validation AP in 12-month pancreatic cancer prediction as the selection criterion. The selected GraphBEAT configuration achieved an AP of $0.541$. Monthly bucketization ($\tau=30$) provided higher AP than the weekly ($\tau=7$) and quarterly ($\tau=90$) alternatives ($0.541$ vs. $0.527$ and $0.502$, respectively). Increasing or decreasing the node and graph feature dimensions from the selected values did not improve performance in this task. The optimal message-passing search depth $K=1$ emphasizes temporally local information in updating feature representations, which aligns with the time-decay edge weighting design. Using three feature pooling heads ($R=3$) yielded the highest validation AP, which may suggest that more enriched feature aggregation across time buckets produces a more informative global graph representation. A mini-batch size of 32 produced similar performance to the selected size of 16 ($0.533$ versus $0.541$), whereas smaller or larger batches led to suboptimal results. 

In addition, among the evaluated hyperparameter settings, the training downsampling ratio produced the largest observed performance variation. The selected ratio of $1{:}3$ achieved an AP of $0.541$, whereas training without downsampling (denoted as original in the table) yielded an AP of $0.184$, underscoring the need for handling data imbalance in this task.

\subsection{Model Explainability Analysis}
Figure~\ref{fig5_explain} summarizes model explanations for 12-month risk prediction across the three cancers on their corresponding MCHS test sets. Because each EHR graph contains large, varying numbers of nodes and edges, 
we focused on features associated with the 10 highest-scoring edges per graph to minimize the potential influence of common or noisy features. 
As shown in the upper panels of Figure~\ref{fig5_explain}, all top-10 contributors to the true-positive predictions in each task are diagnoses, which could be largely related to the high proportion of diagnosis codes in the features. Across the three gastrointestinal conditions, several common codes reflected metabolic disease, chronic disease management, screening, and healthcare interactions, including diabetes (E11), hypertension (I10), long-term drug therapy (Z79), cancer-screening encounters (Z12), immunization encounters (Z23), and musculoskeletal conditions (M25). 

For pancreatic cancer, the edge explanations added pancreas-specific signal. For example, the repeated pancreatic disease records (577$\to$577) ranked among the top influential connections, despite the diagnosis code 577 not appearing among the top-10 individual features. This suggests that the persistence or recurrence of pancreatic conditions was more informative than an isolated diagnosis. Other transitions, including diabetes to long-term drug-therapy coding (E11$\to$Z79) and repeated aftercare encounters following surgery (V58$\to$V58), indicate that the model also emphasized longitudinal patterns in disease and management.

Chronic liver disease (571) was among both the leading features and edges for liver cancer prediction. The transition that involved alcoholic liver disease (K70$\to$Z23) further reinforced the importance of organ-specific disease patterns. Moreover, repeated lung cancer diagnoses (C34$\to$C34) were influential in the liver-cancer classifier, suggesting that the model may also use prior oncologic history for the current liver cancer  prediction. Similar to pancreatic cancer, repeated diabetes-related diagnoses also indicated that persistent metabolic disease contributed to the liver cancer risk prediction.

For intestinal cancer, the highest-ranked features and edges were less organ-specific. Instead, they were primarily dominated by more general codes such as long-term medication use, screening encounters, and chronic disease. This non-specific contribution may partly stem from the broader class definition, which combines small- and large-intestine cancers and could dilute signals specific to a cancer subtype. As such, a broader range of factors, general healthcare interactions, and chronic conditions would contribute more to the model's decision. This could also explain the comparatively poor performance among the three cancers, despite the highest prevalence in our dataset. 

Since the types of laboratory tests were pre-selected based on clinical relevance, the most important blood tests were consistent across all cancers. Specifically, they can be grouped into complete blood count panel (hemoglobin, hematocrit, MCV), chemistry panel (potassium, chloride, and calcium), and metabolic and renal markers (HbA1c and blood urea nitrogen), which were all closely related to the target gastrointestinal conditions. 

The lower panels of Figure~\ref{fig5_explain} show the t-SNE projections~\cite{van2008visualizing} of the graph-level representations after attention-based feature pooling and before classification. Across all three tasks, cancer cases were enriched in localized regions of the 2D projected embedding space rather than being uniformly interspersed with controls. Pancreatic and liver cancer cases formed several compact clusters, whereas intestinal cancer cases occupied a broader but still structured region. Nevertheless, partial case–control separation in the embedding space was observed across all three cancer tasks, which further underscores the heterogeneous and subtle prediagnostic signals that make early cancer risk stratification challenging and clinically invaluable.

\section{Discussion}
In this study, we introduced GraphBEAT, a temporal graph representation learning framework that transforms irregular diagnosis and laboratory histories into patient-specific temporally directed graphs for cancer risk prediction. Across three geographically held-out test sets comprising 107,764 patients, GraphBEAT consistently outperformed the graph-based and sequence modeling baselines in all tasks. At the 12-month lead time, GraphBEAT achieved the mean cross-site AP of 0.523, 0.584, and 0.517 for pancreatic, liver, and intestinal cancer, respectively. Compared with GraphSAGE, the strongest baseline model, GraphBEAT improved AP by $4.8\%$ to $11.4\%$ across different prediction settings. Notably, GraphBEAT used only 12.5M FLOPs on average per patient, compared to 3995.3M for the Transformer and 3230.5M for the LSTM model. Furthermore, with the proposed graph representation, all GNN models outperformed the sequence models while requiring two orders of magnitude lower computation. These findings support a potentially broader clinical application of the our proposed method for modeling heterogeneous longitudinal EHR data.

ICD diagnosis codes are subject to non-clinical misclassification and variation arising from setting-dependent documentation and billing workflows~\cite{hripcsak2013next}, while laboratory measurements may also be affected by within-person physiological and analytical variations~\cite{aarsand2018eubivas}. By grouping irregular clinical events into time buckets and summarizing repeated diagnoses by frequency and laboratory measurements by within-bucket mean, the proposed bucketized encoding method treats events occurring over a short interval as a clinical state rather than as independent temporal tokens. This bucketized representation may therefore attenuate isolated coding errors and short-term measurement variation, while substantially shortening the modeling timeline.  

Converting the bucketized matrix into a layered directed acyclic graph further simplifies the representation by retaining only observed feature-bucket pairs while preserving observation/missingness patterns, measurements, temporal order, and elapsed time between clinical states. Although empty buckets do not generate node layers, the interval with no recorded clinical features is retained through the edge time difference in the message-passing process. Moreover, the three-component node feature representation combines feature identity, observed value, and temporal position in a shared latent space. As such, diagnostic and laboratory nodes, as well as potentially additional data modalities, can exchange information directly through the same topology without requiring separate modality-specific encoders and an additional fusion stage. All the factors collectively contributed to the reduction in computational complexity relative to the sequence baselines.

At the model level, GraphBEAT retains the neighborhood aggregation principle of GraphSAGE but adapts it to the structure of longitudinal EHR data through the proposed context-aware, time-decay message passing. With the selected search depth of $K\text{=1}$, each target node combines its own representation with information from all nodes in the immediately preceding bucket. This one-hop selection focuses message passing on transitions between successive clinical states and would reduce repeated mixing of noisy events, over-smoothing, and unnecessary computation associated with deeper graph propagation. It should also be noted that the one-hop design does not restrict the final prediction to a short clinical interval. In fact, GraphBEAT separates local transition modeling from global evidence aggregation in two steps. Specifically, the message-passing algorithm captures changes between consecutive clinical events, which is consistent with the exponentially time-decayed edge weighting. Then, the multi-head attention feature pooling integrates evidence over the full longitudinal record. Hyperparameter tuning and ablation results further support this EHR-specific modification of conventional neighborhood aggregation.

Although AUROC and AP were used as the primary threshold-independent performance metrics, clinical implementation requires careful selection of the operating threshold to effectively identify high-risk patients based on different clinical requirements. Following prior pancreatic cancer risk stratification work~\cite{varghese2026digitally}, we conducted an exploratory analysis of clinical benefit using a threshold of $0.03$, based on the cost-sensitive learning principle~\cite{elkan2001foundations}, which numerically matched the overall prevalence in our pancreatic cancer cohort. On the MCHS test set, our 12-, 24-, and 36-month models provided positive net benefits of $0.018$, $0.016$, and $0.015$, respectively, translating to $18$, $16$, and $15$ net true-positive cancer detections per 1000 evaluated patients after accounting for the threshold-weighted false-positive penalty. As such, this analysis may suggest a potential use of GraphBEAT with a lower threshold as the first-stage digital enrichment before more specific diagnostic testing. Nevertheless, the final operating threshold should be determined with consideration of model calibration, downstream testing capacity, and clinical requirements.

Several limitations should also be considered. First, all data originated within one healthcare system. Although geographic internal-external validation evaluated transportability across community, tertiary, and quaternary care settings, fully independent validation in external healthcare systems remains necessary. Second, to reflect the ultimate clinical application for disease risk prediction from routine EHR data, cancer cases in this work were defined solely by diagnosis codes, which could contain coding inaccuracies. Further subgroup analysis using verified cases from cancer registries would strengthen the outcome evaluations. Third, downsampling the controls during model training improved performance but also altered class priors, thereby potentially affecting the model's probability calibration. Detailed calibration analysis of the downsampling strategy will be essential in future work before clinical use. Lastly, clinical knowledge-guided feature ablations would be another valuable direction for future work to better understand the model's decision-making for clinical utility.

\section{Conclusion}
In this study, we presented GraphBEAT, a temporal graph representation learning framework that converts irregular diagnosis and laboratory trajectories into patient-specific temporally directed acyclic graphs. By combining three-component node representations, context-aware, time-decay message passing, and attention-based feature pooling, GraphBEAT consistently outperformed graph and sequence modeling baselines across three gastrointestinal cancers, three prediction lead times, and geographically held-out test cohorts, while requiring substantially fewer computations than classic sequence modeling approaches. Experimental results support the proposed temporal graph representation learning framework as an efficient approach for pre-diagnostic risk stratification from irregular, and heterogeneous EHR data. Therefore, the GraphBEAT model holds the potential to serve as a first-stage digital enrichment tool to identify patients in routine clinical records for more specific diagnostic assessment.

\bibliographystyle{IEEEtran}
\bibliography{ref}

\end{document}